\documentclass[square,numbers]{article}

\usepackage[preprint]{neurips_2025}

\usepackage[utf8]{inputenc} % allow utf-8 input
\usepackage[T1]{fontenc}    % use 8-bit T1 fonts
\usepackage{hyperref}       % hyperlinks
\usepackage{url}            % simple URL typesetting
\usepackage{booktabs}       % professional-quality tables
\usepackage{amsfonts}       % blackboard math symbols
\usepackage{nicefrac}       % compact symbols for 1/2, etc.
\usepackage{microtype}      % microtypography
\usepackage{xcolor}         % colors

\usepackage{amsmath}
\usepackage{multirow}
\usepackage{adjustbox}
\usepackage{bbding}
\usepackage{float}    % 提供 [H] 浮动选项
\usepackage{wrapfig}

\title{TopoDiT-3D: Topology-Aware Diffusion Transformer with Bottleneck Structure for 3D Point Cloud Generation}

\author{%
  Zechao Guan\\
  Southeast University\\
  \texttt{zechaoguan@seu.edu.cn} \\
  \And
  Feng Yan\\
  Meituan\\
  \texttt{yanfeng05@meituan.com} \\
  \And
  Shuai Du\\
  Southeast University\\
  \texttt{shuaidu@seu.edu.cn} \\
  \And
  Lin Ma\\
  Meituan\\
  \texttt{malin11@meituan.com} \\
  \And
  Qingshan Liu \thanks{Corresponding Author.} \\
  Southeast University\\
  \texttt{qsliu@seu.edu.cn} \\
}

\begin{document}

\maketitle

\begin{abstract}
Recent advancements in Diffusion Transformer (DiT) models have significantly improved 3D point cloud generation. However, existing methods primarily focus on local feature extraction while overlooking global topological information, such as voids, which are crucial for maintaining shape consistency and capturing complex geometries. To address this limitation, we propose TopoDiT-3D, a Topology-Aware Diffusion Transformer with a bottleneck structure for 3D point cloud generation. Specifically, we design the bottleneck structure utilizing Perceiver Resampler, which not only offers a mode to integrate topological information extracted through persistent homology into feature learning, but also adaptively filters out redundant local features to improve training efficiency. Experimental results demonstrate that TopoDiT-3D outperforms state-of-the-art models in visual quality, diversity, and training efficiency. Furthermore, TopoDiT-3D demonstrates the importance of rich topological information for 3D point cloud generation and its synergy with conventional local feature learning. Videos and code are available at \url{https://github.com/Zechao-Guan/TopoDiT-3D}.
\end{abstract}

\section{Introduction}
As a prevalent representation of 3D data, point clouds and their generation tasks have gained increasing attention due to their various applications, such as shape completion, synthesis, and 3D content creation \citep{hao2021gancraft,huang2019intro,litany2018deformable}. With the growing potential of diffusion transformer (DiT) in high-fidelity image generation \citep{peebles2023DiT-2D,gao2023masked}, extending them to 3D point cloud generation is a promising yet challenging task, due to the unordered nature of point clouds, unlike the structured spatial arrangement of image pixels.

Existing approaches, such as DiT-3D \citep{mo2024dit}, attempt to adapt diffusion transformer to point clouds by transforming them into voxel representations and extracting local features as input tokens through a patch operator. However, this method suffers from two key limitations: (1) \textbf{Redundant tokens} – Due to the sparsity of point clouds, only a small fraction of voxels contain valid data, which leads to most tokens being redundant. This problem worsens as voxel size increases or patch size decreases, further reducing training efficiency and model scalability. (2) \textbf{Lack of global structural awareness} – The patch operator only captures local information, which causes the model to lack global information of point cloud shape. Different from 2D images, point clouds contain a wealth of global information – topological information. However, the challenges lie in extracting the topological information and lacking a method to integrate topological information into feature learning.

\begin{figure}[t]
  \centering
  \includegraphics[width=\textwidth]{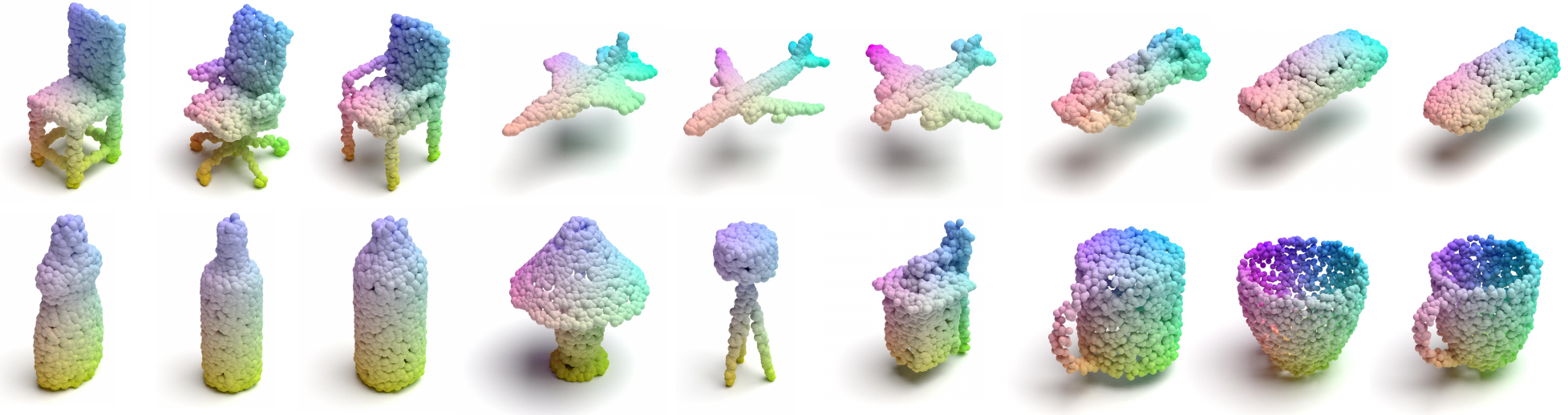}
  \vspace{-0.2cm}
  \caption{Qualitative visualizations demonstrating high-fidelity and diverse 3D point cloud generation.}
  \label{fig:qualitative}
  \vspace{-0.4cm}
\end{figure}

\begin{figure*}[t]
\centering
\includegraphics[width=1\textwidth]{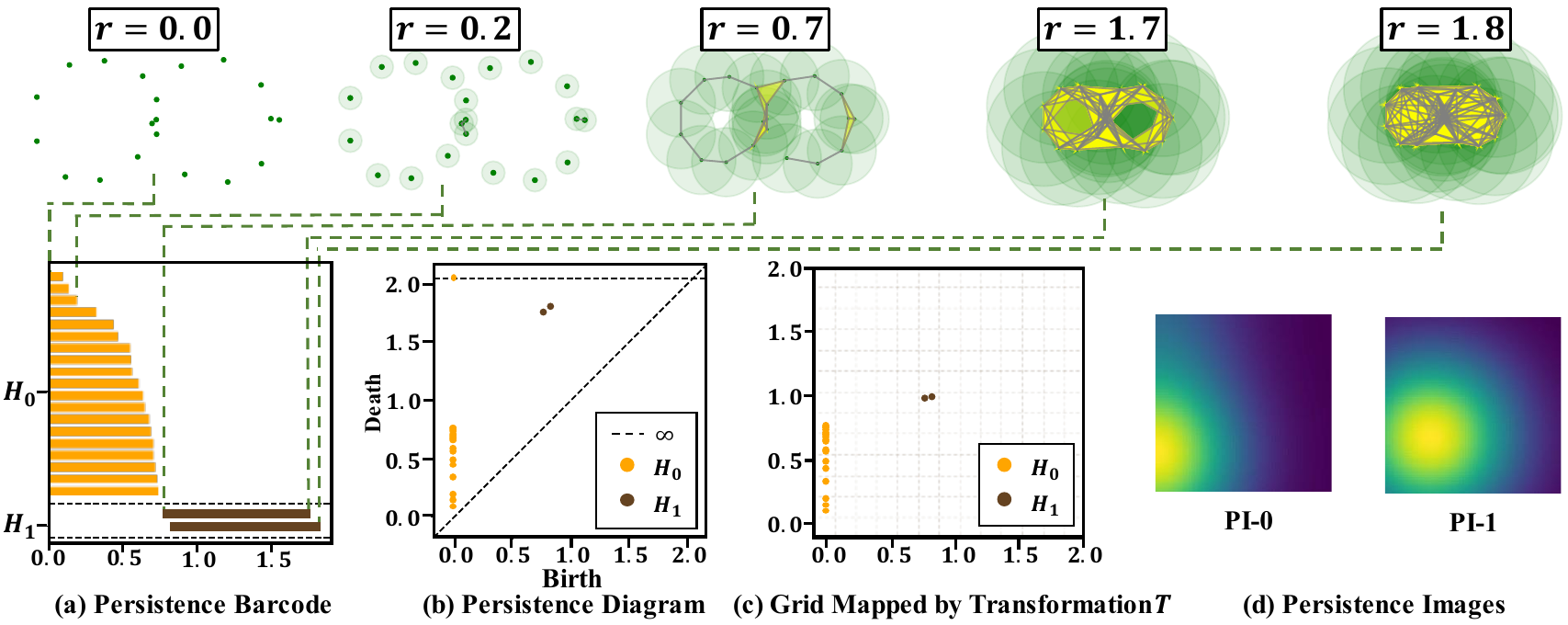} % Reduce the figure size so that it is slightly narrower than the column.
\vspace{-0.2cm}
\caption{The illustration demonstrates the filtration of Vietoris–Rips (VR) complexes, along with the corresponding persistence barcodes, diagrams, and images. As the radius $r$ increases, simplexes are added, leading to the birth and death of topological characteristics. Specifically: (a) Persistence barcodes record the lifespan of homological invariants, with 0-dimension (connected components) in orange and 1-dimension (loops) in brown. (b) Persistence diagram maps birth-death pairs as points, using the same color scheme. (c) and (d) apply a transformation $T$ to map these pairs onto a grid, generating persistence images that encode 0D and 1D topological information.
}
\label{PI-Generation}
\end{figure*}

To tackle these limitations, we propose TopoDiT-3D, a novel diffusion transformer that integrates topological information with an adaptive bottleneck structure to enhance 3D point cloud generation. Our approach utilizes Persistent Homology (PH) from Topological Data Analysis to extract multi-scale topological information, such as connected components, loops, and voids \citep{wasserman2018TDA,cohen2005PD,zomorodian2004computingph,aktas2019phnetwork}. As illustrated in Figure~\ref{PI-Generation}, by constructing a Vietoris-Rips complex and varying the distance threshold, we generate a filtration that encodes topological information into Persistence Diagrams and converts it into Persistence Images, capturing essential global information beyond local geometric features. 

% \begin{wrapfigure}{r}{0.5\textwidth}
%   \centering
%   \includegraphics[width=0.5\textwidth]{figures/training_cost3.pdf}
%   \vspace{-0.2cm}
%   \caption{
%     Comparison of training efficiency between TopoDiT-3D and DiT-3D on an A100 GPU. Experiments were conducted with a batch size of 64, a voxel resolution of 32, and a patch size of 4 over 10,000 training epochs.
%   }
%   \label{train_cost}
% \end{wrapfigure}
\begin{wrapfigure}{r}{0.47\textwidth}
  \vspace{-0.4cm}
  \centering
  \includegraphics[width=0.42\textwidth]{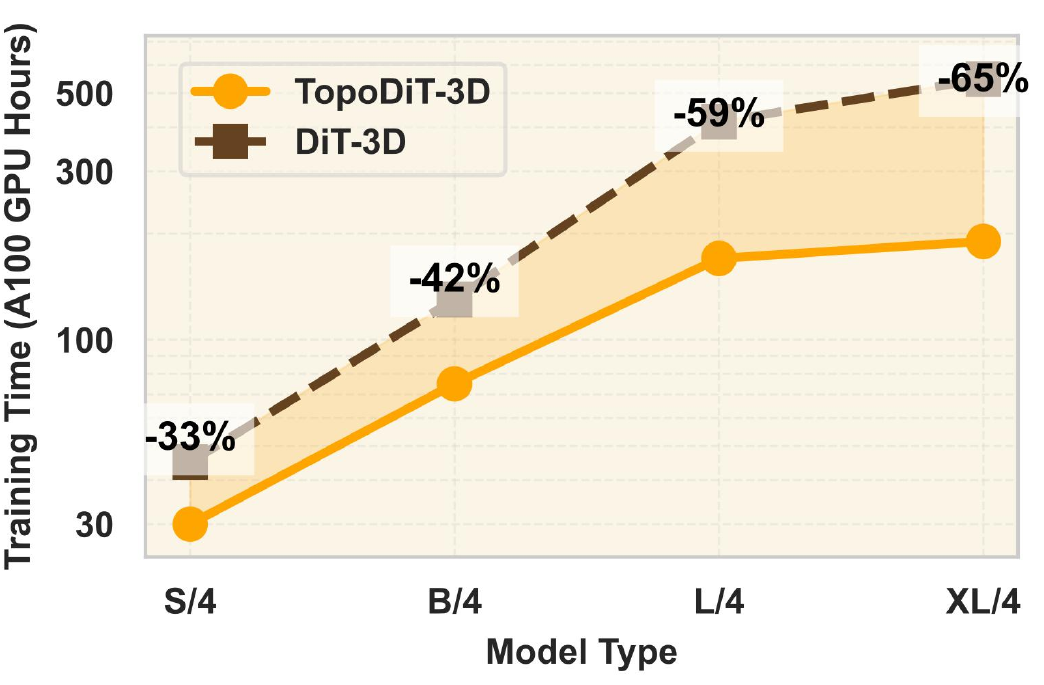}
  \vspace{-0.2cm}
  \caption{Comparison of training efficiency between TopoDiT-3D and DiT-3D. Experiments were conducted with a batch size of 64, a voxel resolution of 32, and a patch size of 4 over 10,000 training epochs.
  }
  \label{train_cost}
  \vspace{-0.2cm}
\end{wrapfigure}

Furthermore, to mitigate token redundancy for improving efficiency, and integrate persistence images into feature learning, we propose a novel bottleneck structure designed with Perceiver Resampler. The Perceiver Resampler is widely used in multi-modal tasks and compresses visual features into the learnable queries \citep{awadalla2023openflamingo,li2023roboflamingo,liu2024robouniview,gong2023multimodal-gpt}. Benefiting from the simple yet effective bottleneck framework, TopoDiT-3D can not only integrate global topological information using persistence images, but also adaptively filter out redundant tokens, and decouple token count from resolution, enhancing the training efficiency and generation capability of the model. As shown in Figure~\ref{train_cost}, when using the XL model size, training time is reduced by 65\%, demonstrating superior efficiency. More importantly, TopoDiT-3D achieves the state-of-the-art performance, surpassing non-DDPM and DDPM baselines on the ShapeNet dataset. The main contributions can be summarized as follows:
\begin{itemize}
    \item We propose TopoDiT-3D, a novel diffusion transformer with a bottleneck structure that uses the Perceiver Resampler to integrate global topological information and filter out redundant tokens, boosting training efficiency;

    \item By using persistent homology, TopoDiT-3D captures multi-scale topological information of point clouds and achieves state-of-the-art performance, surpassing DiT-3D;
    
    \item Our work demonstrates the effectiveness of introducing a bottleneck structure into diffusion transformers and underscores the importance of topological information beyond local features. Extensive experiments on ShapeNet validate the superior performance of TopoDiT-3D over both DDPM and non-DDPM baselines.
\end{itemize}

\section{Related Works}
\subsection{3D Generative Model}
Many prior works have explored point cloud generation in terms of variational autoencoders \citep{yang2018foldingnet,gadelha2018multiresolution,tan2018variational,mittal2022autosdf}, generative adversarial networks \citep{wu2016Intro,yu2020point_encoder_gan,shu20193d,achlioptas2018r-GAN}, and normalized flows \citep{kim2020softflow,pumarola2020C-flow,sanghi2022clip-forge,yang2019pointflow}. Recently, denoising diffusion probabilistic models (DDPMs) have demonstrated remarkable ability across various domains, including image generation \citep{dhariwal2021diffusion,nichol2021glide,zhao2024uni} and 3D point cloud generation \citep{long2024wonder3d,tang2023make,xu2023dream3d}. Several works have explored 3D point cloud generation using DDPMs \citep{luo20213ddiffusion,zhou2021pvdiffusion,vahdat2022lion}. For example, DPM utilized the reverse diffusion process as a Markov chain conditioned on certain point cloud shape latent \citep{luo20213ddiffusion}. As a pioneer, DiT-3D proposed a diffusion transformer architecture specifically designed for 3D point cloud generation \citep{mo2024dit}. However, existing models fail to explicitly model and capture the complex topological structures in point clouds, such as voids, and lack the ability to effectively incorporate additional information for improved feature learning.

%However, existing models fail to explicitly capture the complex topological structures in point clouds, such as voids, and lack the ability to effectively incorporate additional information for improved feature learning.
%In this work, we propose an innovative diffusion transformer architecture with a bottleneck structure utilizing the Perceiver Resampler, which simply and efficiently integrates topological information into feature learning.

\subsection{Persistent Homology (PH) in Deep Learning}
Persistent homology serves as a valuable tool for extracting multi-scale topological information from point clouds, significantly enhancing the effectiveness of point cloud analysis \citep{carlsson2014topological,beksi2016toposegmentation,gabrielsson2020topologylayer}. For instance, the persistent homology based graph convolution network is proposed by \citet{wong2021persistent}, which utilizes PH to extract topological features, thus improving the model for fine-grained 3D shape segmentation. Another work \citep{zhou2022learning} introduced a neural network to learn PH from point clouds, aiming to boost point cloud generation.

In this work, we propose an innovative diffusion transformer bottleneck architecture, TopoDiT-3D, which simply and efficiently integrates topological information into feature learning. Inspired by persistence images \citep{adams2017PI}, which provide a stable vector representation of persistent homology, TopoDiT-3D uses persistence images as its input, enabling the model to capture multi-scale topological information and process point clouds with topological insights.

\section{Preliminaries}
In this section, we initially review the DDPM for 3D point cloud generation. Subsequently, we outline the construction of persistence diagram in the point cloud, setting the preliminaries for persistence image. Detailed definitions of persistent homology can be found in the Appendix.

\subsection{DDPMs on 3D Point Cloud Generation}
Following \citep{zhou2021pvdiffusion,mo2024dit}, we model 3D point cloud generation as a denoising process. In the forward noising process, we define a ground truth diffusion distribution $q(\mathbf{x}_{0:T})$ by gradually applying noise to real data $\mathbf{x}_0$ as $q(\mathbf{x}_t | \mathbf{x}_{t-1})=\mathcal{N} (\mathbf{x}_t ; \sqrt{1- \beta_t } \mathbf{x}_{t-1},\beta_t \mathbf{I})$, where $T$ denotes the number of steps and $q(\mathbf{x}_t | \mathbf{x}_{t-1})$ is a Gaussian transition kernel, which gradually adds noise to the input with a variance schedule $\beta_1, \dots, \beta_T$. The $\beta_T$ are chosen such that the chain approximately converges to a standard Gaussian distribution after $T$ steps. In particular, the denoising process produces a series of shape variables with decreasing levels of noise, denoted as $\mathbf{x}_T, \mathbf{x}_{T-1}, \dots, \mathbf{x}_{0}$, where $\mathbf{x}_T$ is sampled from a Gaussian prior and $\mathbf{x}_{0}$ is the final output. By the reparameterization trick, we have $\mathbf{x}_t = \sqrt{\overline{\alpha}_t} \mathbf{x}_0 + \sqrt{1- \overline{\alpha}_t} \mathbf{\epsilon}$, where $\mathbf{\epsilon} \sim \mathcal{N}(\mathbf{0}, \mathbf{I})$, $\alpha_t = 1 - \beta_t$, and $\overline{\alpha}_t = \prod_{i=1}^t \alpha_i$. In the reverse process, the diffusion model is trained to learn a denoising network $\mathbf{\theta}$, which aims to invert the noise corruption process as $p_\mathbf{\theta} (\mathbf{x}_{t-1} | \mathbf{x}_{t}) = \mathcal{N} (\mathbf{x}_{t-1} ; \mathbf{\mu_\theta} (\mathbf{x}_t, t), \sigma_t^2\mathbf{I})$, where $\mathbf{\mu_\theta} (\mathbf{x}_t, t)$ represents the estimated mean implemented by the network $\mathbf{\theta}$. According to \citet{ho2020denoising}, $\sigma_t$ is a constant, often setting $\sigma_t^2 = \beta_t$ or $\sigma_t^2 = \frac{1-\overline{\alpha}_{t-1}}{1-\overline{\alpha}_t}\beta_t$.

The goal of training the reverse diffusion process is to maximize the log-likelihood: $\mathbb{E}[\log p_\theta(\mathbf{x}_0)]$. However, it is intractable to directly optimize the exact log-likelihood, we instead maximize its variational lower bound:
\begin{align*}
\mathcal{L} = \textstyle -p_\mathbf{\theta}(\mathbf{x}_0 | \mathbf{x}_1) + \sum_{t} \mathcal{D}_\text{KL}(q(\mathbf{x}_{t-1} | \mathbf{x}_t, \mathbf{x}_0) || p_\mathbf{\theta}(\mathbf{x}_{t-1} | \mathbf{x}_t))
\end{align*}
Since both $p_\mathbf{\theta}(\mathbf{x}_{t-1} | \mathbf{x}_t)$ and $q(\mathbf{x}_{t-1} | \mathbf{x}_t, \mathbf{x}_0)$ are Gaussian, $\mathbf{\mu_\theta} (\mathbf{x}_t, t)$ can be reparameterized using the predicted noise $\mathbf{\epsilon_\theta}(\mathbf{x}_t, t)$. Finally, the loss can be reduced to a mean squared error (MSE) loss $\mathcal{L}_2$ between the model's predicted noise $\mathbf{\epsilon_\theta}(\mathbf{x}_t, t)$ and the ground truth Gaussian noise $\mathbf{\epsilon}$. After the model is trained, we can generate point clouds by progressively sampling from $p_\mathbf{\theta}(\mathbf{x}_{t-1} | \mathbf{x}_t)$ as $t = T, \dots, 1$.

\subsection{Persistent Homology}

\begin{figure*}[t]
\centering
\includegraphics[width=1\textwidth]{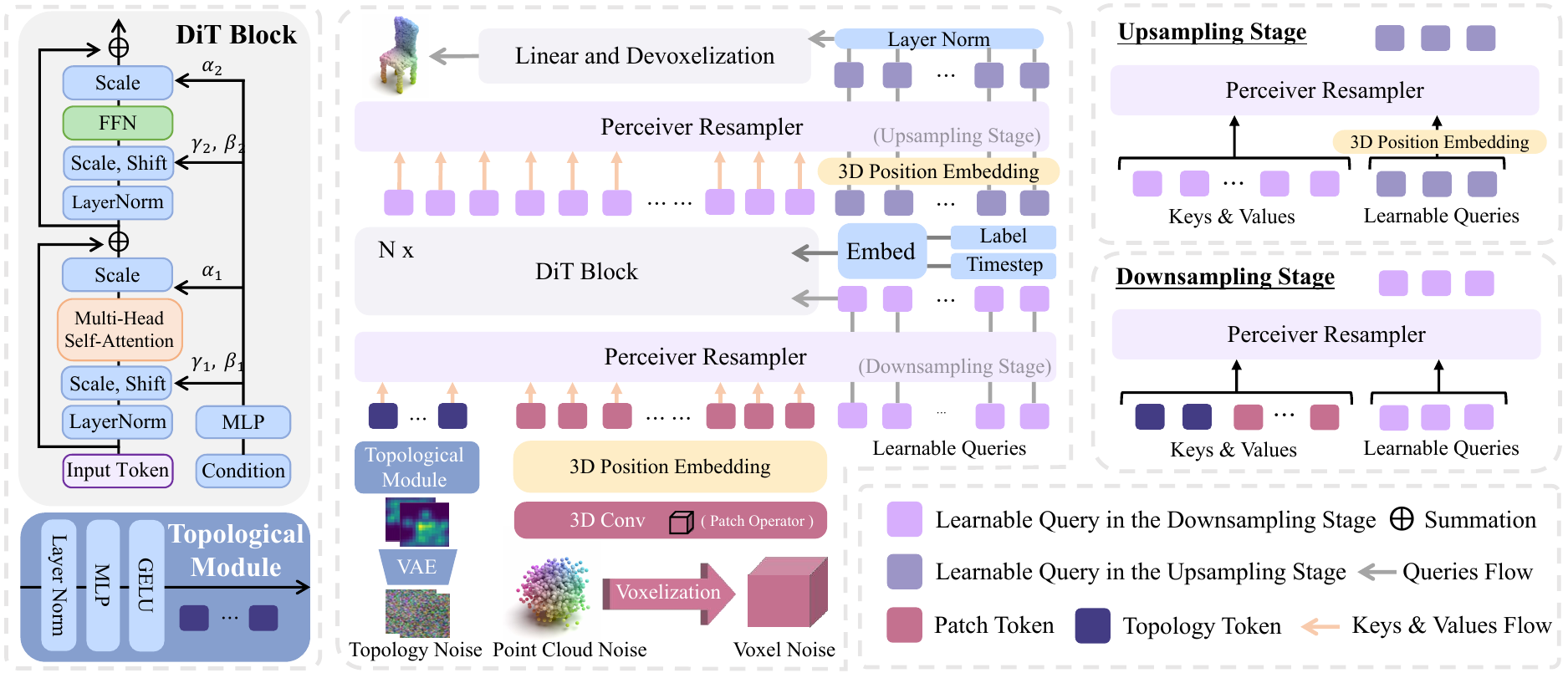} % Reduce the figure size so that it is slightly narrower than the column.
\vspace{-0.2cm}
\caption{The illustration of the proposed Topology-Aware Diffusion Transformer (TopoDiT-3D) for 3D point cloud generation. TopoDiT-3D initially voxelizes the point clouds, employing the patch operator to generate tokens related to the local point-voxel feature, and the persistence images to generate tokens related to the global topological feature. The persistence images are generated by the pretrained VAE during inference. Subsequently, TopoDiT-3D uses a fixed minor number of learned queries and the Perceiver Resampler to downsample and learn the topological and geometric information. After $N$ DiT blocks, it uses the Perceiver Resampler to achieve upsampling, which recovers the same number of patch tokens to devoxelize.}
\label{architecture}
\end{figure*}

Persistent homology is a powerful tool for identifying and describing topological features within datasets, such as connected components, loops, voids, and more. These features can be represented by the Persistence Diagram (PD), which provides a comprehensive view of the topological information. The PD is generated through three key processes:

\paragraph{Building the Vietoris-Rips Complex.}
Given a point cloud $P$ consisting of points $x_i\textit{ }(i \in \{1, \dots, N\})$, we define a radius $r$ as a distance threshold for establishing connectivity within the point cloud $P$. Specifically, a point $x_j$ is considered connected to point $x_i$ if $x_j$ lies within a ball centered at $x_i$ with radius $r$. A $k$-simplex is formed when $k+1$ points are mutually connected in this manner \citep{chambers2010Vietoris-rips}. The set of these $k$-simplexes with varying dimensions $k$, collectively forms the Vietoris-Rips complex.

\paragraph{Generating the Filtration and utilizing Persistent Homology.}
As the radius $r$ increases from zero demonstrated in Figure~\ref{PI-Generation}(a), new $k$-simplexes are formed, and higher-dimensional $k^\prime$-simplexes ($k^\prime > k$) may appear as lower-dimensional ones become connected. This process, called filtration, generates a nested sequence of complexes from the point cloud \citep{zhou2022learning}. Persistent homology tracks the birth and death of $k$-dimensional homological invariants during filtration. These invariants, which represent topological characteristics such as connected components, loops, and voids, correspond to $k$ values of 0, 1, and 2, respectively.

\paragraph{Persistence Diagram.}
During filtration, the topological information extracted from the point cloud are summarized in the Persistence Diagram (PD). As illustrated in Figure~\ref{PI-Generation}(b), the PD is a set of points in $R^{2}$. For each fixed homological dimension $k$, the $k$-dimensional homological invariants are represented as points $(x,y)$ in the diagram, where $x$ is the birth time of the invariant with respect to the radius $r$, and $y$ is the death time. The difference $y - x$ referred to as persistence, quantifies the significance of the invariant.

\section{Methods}
In this section, we propose TopoDiT-3D, a novel topology-aware diffusion transformer with bottleneck structure, as shown in Figure~\ref{architecture}. We first overview the TopoDiT-3D architecture, followed by the construction of persistence images to extract topology tokens. Finally, we elaborate on the bottleneck structure, which leverages the Perceiver Resampler.

\subsection{TopoDiT-3D Architecture Overview}
For each point cloud $P_{i} \in R^{N \times 3}$, where $N$ represents the number of points with $x,y,z$ coordinates, TopoDiT-3D initially voxelizes it as input $v_{i} \in R^{V \times V \times V \times 3}$, where $V$ represents the voxel size. For local geometric information, following the DiT-3D \citep{mo2024dit}, we then process this voxel using a patchification operator, dividing it into smaller patches of size $p \times p \times p$, resulting in a sequence of patch tokens $s_{i} \in R^{L \times d}$ with length $L=\left(V/p\right)^3$. For global topological information, we construct persistence images to extract topology tokens $t_{i} \in R^{2 \times d}$. These tokens pass through a bottleneck structure with multiple transformer blocks (DiT block) that employ self-attention to update the features. Finally, to achieve denoising in the point cloud space, the voxel is converted back into the original point cloud format by devoxelization, resulting in the output noise $\epsilon_{\theta}\left(x_{t}, t\right)$.

\subsection{Topological Information}
For ease of representation, the persistence diagram (PD) is denoted as the multi-set $B$ containing points $(x_{i}, y_{i})$, where $x_{i}$ represents the birth and $y_{i}$ denotes the death of topological characteristics. As a stable vector representation \citep{adams2017PI}, the persistence image maps the PD into a two-dimensional grid, effectively transforming the topological information into a structured pixel representation through three key processes:

\paragraph{Mapping to a grid.}
As depicted by Figure~\ref{PI-Generation}(c), utilizing the linear transformation $T: R^{2} \rightarrow R^{2}$ defined by $T(x,y)=(x,y-x)$, each point $(x_{i}, y_{i})$ is mapped to a pixel location in a two-dimensional grid, denoted as $T(B)$, considering the persistence values as coordinates.

\paragraph{Constructing the persistence surface.}
To construct a differentiable persistence surface, a differentiable Gaussian probability distribution $\phi_{u}:R^{2}\rightarrow R$ is employed. It is parameterized by the mean $u = (u_{x}, u_{y}) \in R^{2}$ and a non-negative weighting function $f:R^{2}\rightarrow R$. The persistence surface is expressed as:
\begin{align}
\rho_{B}(x,y) = \textstyle \sum_{u \in T(B)}f(u)\phi_{u}(x,y)
\end{align}
where the commonly chosen differentiable Gaussian function is given by $\phi_{u} = g_{u}(x,y;\sigma)$$ = \frac{1}{2\pi\sigma^{2}}e^{-\frac{(x-u_{x})^2+(y-u_{y})^2}{\sigma^{2}}}$, where the $\sigma$ often set to $1$. The weighting function is $f=\frac{y-x}{b}$, where $b$ represents the largest persistence value among all persistence diagrams in the point clouds.

\paragraph{Creating persistence image.}

A grid in the plane with $n$ pixels is defined, and each pixel $p$ is assigned the integral of $\rho _{B}$ over that region, expressed as:
\begin{align}
I(\rho_{B})_{p} = \iint_{p}{\rho_{B}}\mathrm{d}y\mathrm{d}x
\end{align}
where the Figure~\ref{PI-Generation}(d) shows the 0D and 1D persistence images (PI-0 and PI-1). In point cloud generation, we use PI-1 and PI-2, focusing on structures such as loops and voids. 

% Our TopoDiT-3D captures local information through the patchification operator, while the persistence images capture the overall shape and structural information of point cloud as global topological information. This allows for multi-scale feature fusion, resulting in a more comprehensive and rich feature representation. To fully capture the topological information embedded in the persistence images (PIs), we flatten the PIs and input them into the Topological module, which consists of LayerNorm, MLP, and GELU activations, ultimately producing the Topology tokens $t_{i} \in R^{2 \times d}$. Additionally, to utilize these global features during the denoising process in inference, we pretrain a VAE model \citep{kingma2013auto} to generate the persistence images, thus guiding the model to generate more accurate point clouds.
TopoDiT-3D captures local features via a patchification operator and global topological structures through persistence images (PIs), enabling multi-scale feature fusion. To fully capture the topological information embedded in the PIs, we flatten the PIs and input them into a lightweight topological module (consisting of LayerNorm, MLP, GELU), producing topology tokens $t_i \in \mathbb{R}^{2 \times d}$. However, noisy point clouds lack clear topological structure, making it difficult to extract reliable PIs during inference. To address this, we pretrain a VAE \cite{kingma2013auto} to generate PIs as global priors, which are used during the denoising process at inference time to guide the model in producing structurally coherent and topologically faithful point clouds. Details of the topological module and VAE architecture are provided in the Appendix.

\subsection{Bottleneck Structure}

\begin{wrapfigure}{r}{0.3\textwidth}
\vspace{-0.6cm}
\centering
\includegraphics[width=0.3\textwidth]{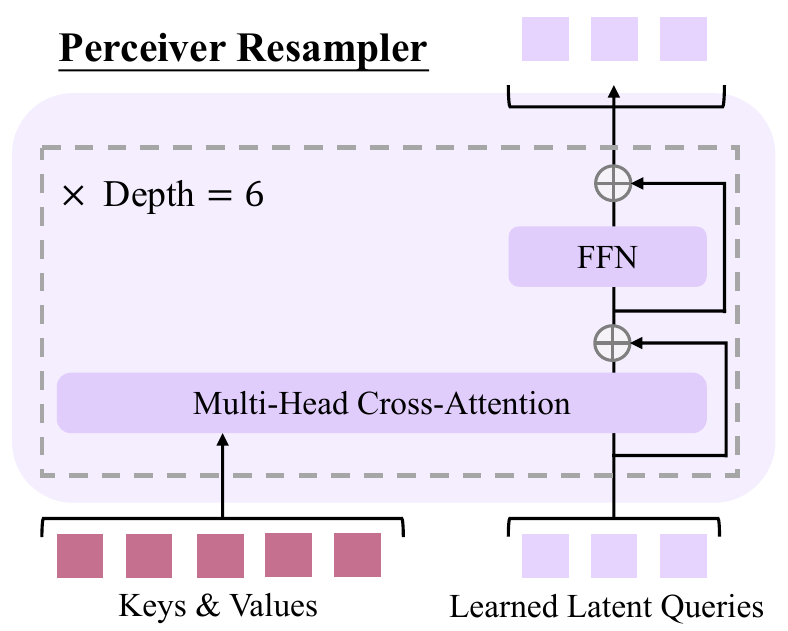} % Reduce the figure size so that it is slightly narrower than the column.
\vspace{-0.5cm}
\caption{The Perceiver Resampler module.}
\label{perceiver_resampler_architecture}
\vspace{-0.5cm}
\end{wrapfigure}

DiT-3D transforms voxels into patch tokens via the patchification operator, but this process has some limitations. Firstly, due to point cloud sparsity, only a minor portion of voxels carry valid data. Secondly, an increase in resolution leads to a surge in patch tokens, negatively impacting model scalability and reducing training efficiency. Thirdly, DiT-3D lacks a method to integrate information from other modalities into feature learning.

To overcome these issues, we employ the Perceiver Resampler to design a simple but efficient bottleneck structure. This structure can adaptively filter out redundant and information-less patch tokens, decoupling the number of tokens entering the DiT block from resolution, thus enhancing training efficiency. Importantly, it allows the incorporation of topological information into feature learning. As shown in Figure~\ref{perceiver_resampler_architecture}, the Perceiver Resampler can be defined by:
\begin{align}
l_{j}^{resample} = \text{MHCA}\left(k, v, q_{j}^{resample}\right) + q_{j}^{resample}\\
q_{j+1}^{resample} = \text{FFN}\left(l_{j}^{resample}\right) + l_{j}^{resample}
\end{align}
where MHCA represents the Multi-Head Cross-Attention \citep{vaswani2017attention}, which receives a predefined number $M$ of learnable latent input queries $q_{j}^{resample} \in R^{M \times d}$, cross-attend to the features $k$ and $v$, and then inputs to the Feed-Forward Network (FFN), with $j=1,...,\text{depth}$. As illustrated in Figure~\ref{architecture}, the bottleneck structure consists of the following two stages:

\paragraph{Downsampling stage.}
During the downsampling phase, we utilize $M=96$ learnable latent queries $q^{down}$, and $k, v$ represent the patch tokens $s_{i} \in R^{L \times d}$ and the topology tokens $t_{i} \in R^{2 \times d}$. These learned latent queries can adaptively filter out redundant, information-less patch tokens, and learn from global topological information.

\paragraph{Upsampling stage.}
During the upsampling phase, we utilize $M=L$ learnable latent queries $q^{up}$ to restore the same token number as $s_{i}$ for devoxelization, and $k, v$ represent the $q^{down}$ updated from the diffusion transformer. Notably, $q^{up}$ are added with the 3D position embedding identical to $s_{i}$, which assists the model in understanding the placement of each token in the reconstructed space.

Similar to DiT-3D \citep{mo2024dit} and DiT \citep{peebles2023DiT-2D}, TopoDiT-3D is a scalable model capable of accommodating various voxel and patch sizes, as well as different model scales. It can flexibly adapt to voxel sizes of 16, 32, 64, patch sizes of 2, 4, 8, and model complexities including Small, Base, Large, and Extra Large, as depicted in the DiT and DiT-3D model. For instance, the TopoDiT-3D-S/4 model signifies the usage of the small configuration of the DiT and DiT-3D models with a patch size of 4.

\section{Experiments}
In this section, we comprehensively evaluate TopoDiT-3D and conduct in-depth ablation experiments to investigate the effectiveness of the bottleneck structure and topological information within point cloud data.

\subsection{Experiment Setup}
\paragraph{Datasets.}
In line with most of the previous works \citep{yang2019pointflow,luo20213ddiffusion,zhou2021pvdiffusion,vahdat2022lion,mo2024dit}, we use ShapeNet dataset as the benchmark for 3D point cloud generation. The ShapeNet benchmark comprises a dataset of 51,127 shapes categorized into 55 classes. Following the previous works, we train TopoDiT-3D on three categories: Chair, Airplane and Car. For a fair comparison, we also maintain PointFlow's \citep{yang2019pointflow} preprocessing and dataset splits, which normalizes the data globally across the whole dataset and randomly selects 2048 points for training and testing.

\paragraph{Evaluation.}

As discussed in PointFlow \citet{yang2019pointflow}, there are numerous evaluation metrics for point cloud generation models, focusing on different aspects. Following the previous works \citep{vahdat2022lion,mo2024dit}, we use Chamfer Distance (CD) and Earth Movers Distance (EMD) as distance metrics to calculate 1-Nearest Neighbor Accuracy (1-NNA) and Coverage (COV) as evaluation metrics. 1-NNA represents the leave-one-out accuracy of the 1-Nearest Neighbor classifier, where a lower 1-NNA score indicates better model performance in terms of quality and diversity of the generated point clouds. COV measures the number of reference point clouds matched to at least one generated shape. While it effectively quantifies diversity and is sensitive to mode dropping, it does not measure the quality of the generated point clouds.

\paragraph{Implementation.}
%To ensure fairness, we adopt the implementation from DiT-3D for all tasks \cite{mo2024dit}. We set the input voxel size as $32 \times 32 \times 32 \times 3$ ($V = 32$) and train the models for 10,000 epochs using the Adam optimizer with a learning rate of $1\times10^{-4}$ and a batch size of $512$. For our experiments, we set $T = 1000$ and use XL/4 with a patch size of $p = 4$ as the default backbone. Specifically, we set the depth of Perceiver Resampler to 6, with 96 tokens for downsampling and 512 tokens for upsampling. We use Farthest Point Sampling to select a subset of key points ($N_{\text{PD}}=64$) from the original point cloud to reduce the computational complexity of calculating persistence diagrams while preserving shape homology \cite{levrard2022distance}. For the VAE that generates persistence images (PI-1 and PI-2), we train for 1,000 epochs using the Adam optimizer with a learning rate of $5\times10^{-3}$ and a batch size of 64.

To ensure consistency across tasks, we adopt the implementation from DiT-3D \citep{mo2024dit}. The input voxel size is set to $32 \times 32 \times 32 \times 3$ ($V = 32$), and we train the models for 10,000 epochs using the Adam optimizer with a learning rate of $1\times10^{-4}$ and a batch size of 512. In our experiments, we set $T = 1000$ and use the XL/4 configuration with a patch size of $p = 4$ as the default backbone. Specifically, we set the depth of Perceiver Resampler to 6, with 96 tokens for downsampling and 512 tokens for upsampling. We use Farthest Point Sampling to select a subset of key points ($N_{\text{PD}}=64$) from the original point cloud to reduce the computational complexity of calculating persistence diagrams while preserving shape homology \citep{levrard2022distance}. For the VAE generating persistence images (PI-1 and PI-2), we train for 1,000 epochs using the Adam optimizer with a learning rate of $5\times10^{-3}$ and a batch size of 64.

\subsection{Comparison to State-of-the-art Works}
\begin{table*}[t]
\caption{Experimental results compare TopoDiT-3D with state-of-the-art methods on 3D point cloud generation across chair, airplane, and car categories.}
\label{tab_three_categories}
\begin{adjustbox}{width=\textwidth,center}
\begin{tabular}{l|llll|llll|llll}
\hline
& \multicolumn{4}{c|}{Chair} & \multicolumn{4}{c|}{Airplane} & \multicolumn{4}{c}{Car} \\
& \multicolumn{2}{c}{1-NNA{\color[HTML]{CB0000}$(\downarrow )$}} & \multicolumn{2}{c|}{COV{\color[HTML]{32CB00}$(\uparrow )$}} & \multicolumn{2}{c}{1-NNA{\color[HTML]{CB0000}$(\downarrow )$}} & \multicolumn{2}{c|}{COV{\color[HTML]{32CB00}$(\uparrow )$}} & \multicolumn{2}{c}{1-NNA{\color[HTML]{CB0000}$(\downarrow )$}} & \multicolumn{2}{c}{COV{\color[HTML]{32CB00}$(\uparrow )$}} \\
\multirow{-3}{*}{Method} & CD & EMD & CD & EMD & CD & EMD & CD & EMD & CD & EMD & CD & EMD \\ \hline
r-GAN \cite{achlioptas2018r-GAN} & 83.69 & 99.70 & 24.27 & 15.13 & 98.40 & 96.79 & 30.12 & 14.32 & 94.46 & 99.01 & 19.03 & 6.539 \\
l-GAN (CD) & 68.58 & 83.84 & 41.99 & 29.31 & 87.30 & 93.95 & 38.52 & 21.23 & 66.49 & 88.78 & 38.92 & 23.58 \\
l-GAN (EMD) & 71.90 & 64.65 & 38.07 & 44.86 & 89.49 & 76.91 & 38.27 & 38.52 & 71.16 & 66.19 & 37.78 & 45.17 \\
PointFlow \cite{yang2019pointflow}& 62.84 & 60.57 & 42.90 & 50.00 & 75.68 & 70.74 & 47.90 & 46.41 & 58.10 & 56.25 & 46.88 & 50.00 \\
SoftFlow \cite{kim2020softflow} & 59.21 & 60.05 & 41.39 & 47.43 & 76.05 & 65.80 & 46.91 & 47.90 & 64.77 & 60.09 & 42.90 & 44.60 \\
SetVAE \cite{kim2021setvae} & 58.84 & 60.57 & 46.83 & 44.26 & 76.54 & 67.65 & 43.70 & 48.40 & 59.94 & 59.94 & 49.15 & 46.59 \\
DPF-Net \cite{klokov2020DPF-Net} & 62.00 & 58.53 & 44.71 & 48.79 & 75.18 & 65.55 & 46.17 & 48.89 & 62.35 & 54.48 & 45.74 & 49.43 \\ \hline
DPM \cite{luo20213ddiffusion} & 60.05 & 74.77 & 44.86 & 35.50 & 76.42 & 86.91 & 48.64 & 33.83 & 68.89 & 79.97 & 44.03 & 34.94 \\
PVD \cite{zhou2021pvdiffusion} & 57.09 & 60.87 & 36.68 & 49.24 & 73.82 & 64.81 & 48.88 & 52.09 & 54.55 & 53.83 & 41.19 & 50.56 \\
LION \cite{vahdat2022lion} & 53.70 & 52.34 & 48.94 & 52.11 & 67.41 & 61.23 & 47.16 & 49.63 & 53.41 & 51.14 & 50.00 & 56.53 \\ \hline
GET3D \cite{gao2022get3d} & 75.26 & 72.49 & 43.36 & 42.77 & - & - & - & - & 75.26 & 72.49 & 15.04 & 18.38 \\
MeshDiffusion \cite{liu2023meshdiffusion}& 53.69 & 57.63 & 46.00 & 46.71 & 66.44 & 76.26 & 47.34 & 42.15 & 81.43 & 87.84 & 34.07 & 25.85 \\ \hline
DiT-3D \cite{mo2024dit} & 49.11 & 50.73 & 52.45 & 54.32 & 62.35 & 58.67 & 53.16 & 54.39 & 48.24 & 49.35 & 50.00 & 56.38 \\
TopoDiT-3D (ours) & \textbf{46.91}{\color[HTML]{CB0000}(-2.2)} & \textbf{46.43}{\color[HTML]{CB0000}(-4.3)} & \textbf{54.51}{\color[HTML]{32CB00}(+2.06)} & \textbf{56.62}{\color[HTML]{32CB00}(+2.3)} & \textbf{53.18} {\color[HTML]{CB0000}(-9.17)} & \textbf{50.61}{\color[HTML]{CB0000}(-8.06)} & \textbf{64.70}{\color[HTML]{32CB00}(+11.54)} & \textbf{61.27}{\color[HTML]{32CB00}(+6.88)} & \textbf{44.46}{\color[HTML]{CB0000}(-3.78)} & \textbf{43.46}{\color[HTML]{CB0000}(-5.89)} & \textbf{55.56}{\color[HTML]{32CB00}(+5.56)} & \textbf{62.50}{\color[HTML]{32CB00}(+6.12)} \\ \hline
\end{tabular}
\end{adjustbox}
\end{table*}

\begin{figure}[t]
\centering
\begin{minipage}{0.475\linewidth}
    \centering
    \includegraphics[width=\linewidth]{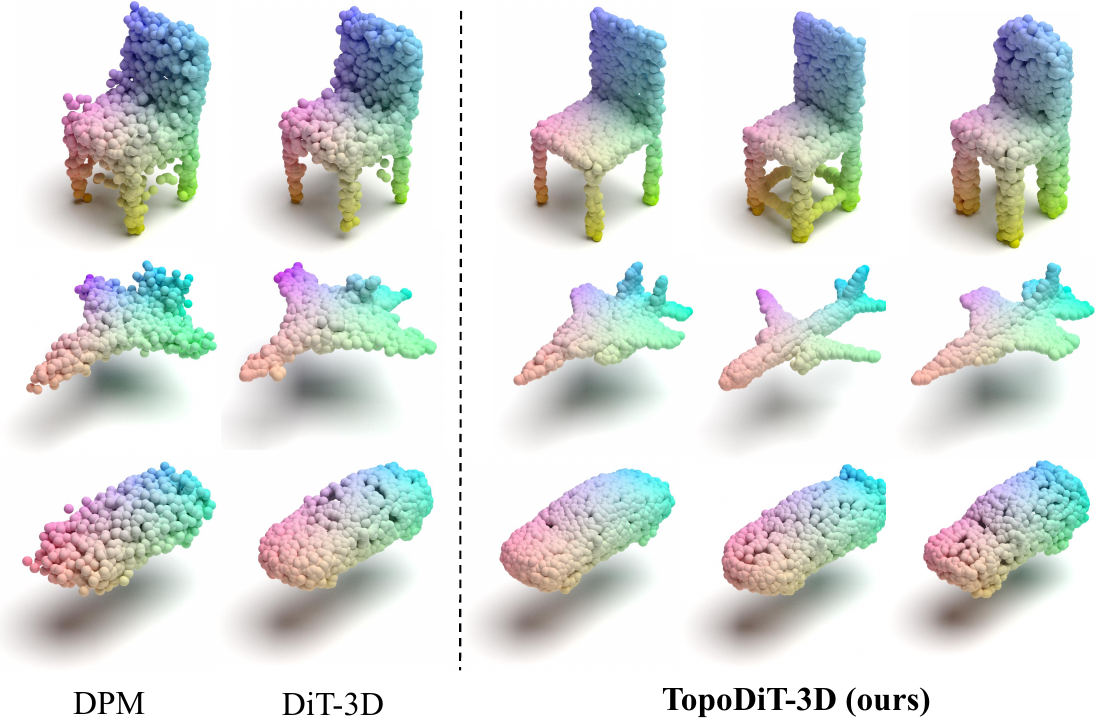}
    \vspace{-0.4cm}
    \caption{The visualizations of 3D point cloud generation.}
    \label{visual}
    \vspace{-0.2cm}
\end{minipage}
\hfill
\begin{minipage}{0.5\linewidth}
    \centering
    \includegraphics[width=\linewidth]{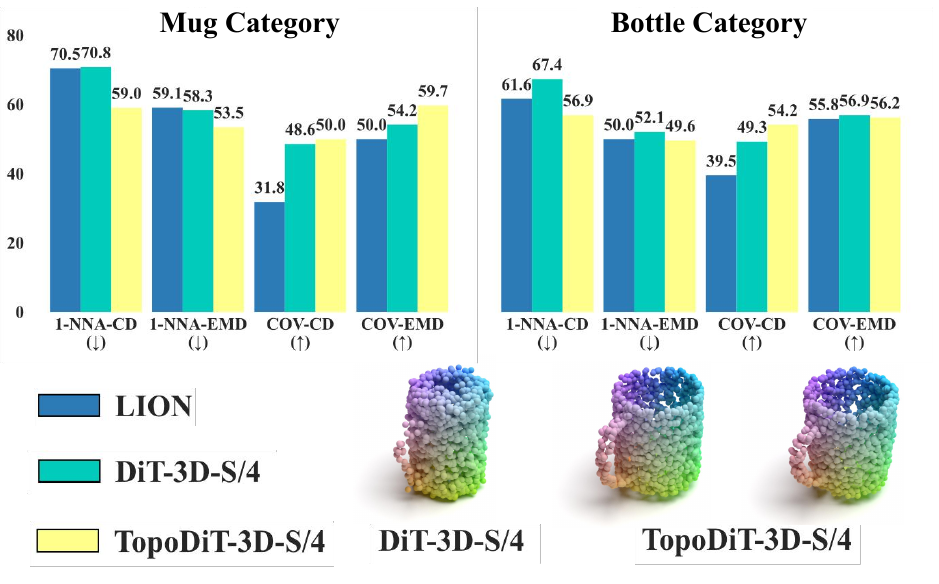}
    \vspace{-0.4cm}
    \caption{The result of ShapeNet’s mug and bottle categories.}
    \label{mug_bottle}
    \vspace{-0.2cm}
\end{minipage}
\end{figure}

We propose TopoDiT-3D, an innovative diffusion transformer with a bottleneck, which can also incorporate topological information as global features. It outperforms the state-of-the-art methods in both effectiveness and training efficiency. To demonstrate the powerful performance of TopoDiT-3D, we compare it with various benchmark models, including GANs \citep{achlioptas2018r-GAN}, Flows \citep{yang2019pointflow,kim2020softflow,klokov2020DPF-Net}, VAEs \citep{kim2021setvae}, Diffusion Probabilistic Models \citep{luo20213ddiffusion,zhou2021pvdiffusion,vahdat2022lion}, and Diffusion Transformer-based \citep{mo2024dit} approaches. 

The quantitative results are shown in Table~\ref{tab_three_categories}. In 3D point cloud generation, TopoDiT-3D significantly improves both the 1-NNA and COV metrics for the chair, airplane, and car categories, outperforming the state-of-the-art model DiT-3D. For the chair category, TopoDiT-3D reduces the 1-NNA of CD and EMD by 2.2 and 4.3, respectively, while boosting the COV by 2.06 and 2.3. Specifically, for the airplane category, which contains rich topological information, TopoDiT-3D reduces the 1-NNA of CD and EMD by 9.17 and 8.06, respectively, and increases the COV by 11.54 and 6.88. These results demonstrate that TopoDiT-3D outperforms non-DDPM and DDPM baselines in terms of generation quality and diversity, showcasing its superior performance. The visual results in Figure~\ref{visual} further highlight its ability to generate high-quality and diverse 3D point clouds.

\subsection{Ablation Analysis}
In this section, we conduct detailed ablation experiments to explore the impact of the bottleneck structure and topological information. Additionally, we evaluate the generation capability across different categories, resolutions, and the scalability of TopoDiT-3D. More experiments, visualization, and further analysis can be found in the Appendix.

\paragraph{Ablation studies on the components of TopoDiT-3D.}
Table~\ref{tab_model_components_ablation} summarizes the performance of TopoDiT-3D under different component configurations. Compared to the base model DiT-3D-S/4, integrating all components in TopoDiT-3D-S/4 improves COV metric by 11.07 (CD) and 9.94 (EMD) while reducing 1-NNA metric by 6.47 (CD) and 9.36 (EMD), achieving performance comparable to DiT-3D-XL/4 with only 30 A100 hours for training.

\begin{wraptable}{r}{0.5\textwidth}
\vspace{-0.6cm}
\caption{Ablation studies on the components of TopoDiT-3D for the chair generation task. Notably, the position embedding refers to whether or not it is added in the upsampling stage.}
\label{tab_model_components_ablation}
\begin{adjustbox}{width=0.5\textwidth,center}
\begin{tabular}{ccccllll}
\hline
\multirow{2}{*}{Model} & \multirow{2}{*}{\begin{tabular}[c]{@{}c@{}}Bottleneck \\ Structure\end{tabular}} & \multirow{2}{*}{\begin{tabular}[c]{@{}c@{}}Position \\ Emdding\end{tabular}} & \multirow{2}{*}{\begin{tabular}[c]{@{}c@{}}Topological \\ Information\end{tabular}} & \multicolumn{2}{c}{1-NNA$(\downarrow)$} & \multicolumn{2}{c}{COV$(\uparrow)$} \\
& & &  & CD & EMD & CD & EMD \\ \hline

\multicolumn{1}{l}{DiT-3D-S/4} & \color[HTML]{CB0000}\XSolidBold & \color[HTML]{CB0000}\XSolidBold & \color[HTML]{CB0000}\XSolidBold & 56.31 & 55.82 & 47.21 & 50.75 \\
\hline

\multirow{4}{*}{\begin{tabular}[c]{@{}c@{}}\\ \\ TopoDiT-3D-S/4\end{tabular}}  

& \color[HTML]{32CB00}\CheckmarkBold & \color[HTML]{CB0000}\XSolidBold & \color[HTML]{CB0000}\XSolidBold & \begin{tabular}[c]{@{}l@{}}57.76\\ (+1.45)\end{tabular} & \begin{tabular}[c]{@{}l@{}}59.39\\ (+3.57)\end{tabular} & \begin{tabular}[c]{@{}l@{}}49.80\\ (+2.59)\end{tabular} & \begin{tabular}[c]{@{}l@{}}48.75\\ (-2.00)\end{tabular} \\

& \color[HTML]{CB0000}\XSolidBold & \color[HTML]{32CB00}\CheckmarkBold & \color[HTML]{CB0000}\XSolidBold & \begin{tabular}[c]{@{}l@{}}57.85\\ (+1.54)\end{tabular} & \begin{tabular}[c]{@{}l@{}}56.65\\ (+0.83)\end{tabular} & \begin{tabular}[c]{@{}l@{}}51.85\\ (+4.64)\end{tabular} & \begin{tabular}[c]{@{}l@{}}51.65\\ (+0.90)\end{tabular} \\

& \color[HTML]{32CB00}\CheckmarkBold & \color[HTML]{32CB00}\CheckmarkBold & \color[HTML]{CB0000}\XSolidBold & \begin{tabular}[c]{@{}l@{}}52.86\\ (-3.45)\end{tabular} & \begin{tabular}[c]{@{}l@{}}54.84\\ (-0.98)\end{tabular} & \begin{tabular}[c]{@{}l@{}}52.56\\ (+5.35)\end{tabular} & \begin{tabular}[c]{@{}l@{}}51.05\\ (+0.30)\end{tabular} \\

& \color[HTML]{32CB00}\CheckmarkBold & \color[HTML]{32CB00}\CheckmarkBold & \color[HTML]{32CB00}\CheckmarkBold & \begin{tabular}[c]{@{}l@{}}\textbf{49.84}\\ (-6.47)\end{tabular} & \begin{tabular}[c]{@{}l@{}}\textbf{46.46}\\ (-9.36)\end{tabular} & \begin{tabular}[c]{@{}l@{}}\textbf{58.28}\\ (+11.07)\end{tabular} & \begin{tabular}[c]{@{}l@{}}\textbf{60.69}\\ (+9.94)\end{tabular} \\ \hline
\end{tabular}
\end{adjustbox}
\vspace{-0.6cm}
\end{wraptable}

Notably, performance drops when only using the bottleneck structure or applying position embedding for patch tokens after the DiT block. Without extra spatial cues, the Perceiver Resampler during upsampling struggles to capture fine-grained structures, while position embedding becomes redundant for patch tokens. By contrast, the inclusion of position embedding, which provides local spatial cues, and topological information, which offers a global view of the topological structure, enables the TopoDiT-3D model to better understand and process point clouds at different scales, leading to an overall performance boost.

% \begin{figure}[htbp]
% \centering
% \begin{minipage}{0.8\textwidth}
%     \centering
%     \includegraphics[width=1\textwidth]{figures/topology_impact_horizon.pdf} % Reduce the figure size so that it is slightly narrower than the column.
% \end{minipage}%
% \hfill
% \begin{minipage}{0.2\textwidth}
% \caption{The impact of topological information.}
% \label{generative_process}
% \end{minipage}
% \end{figure}

\begin{figure}[t]
\centering
\includegraphics[width=1\textwidth]{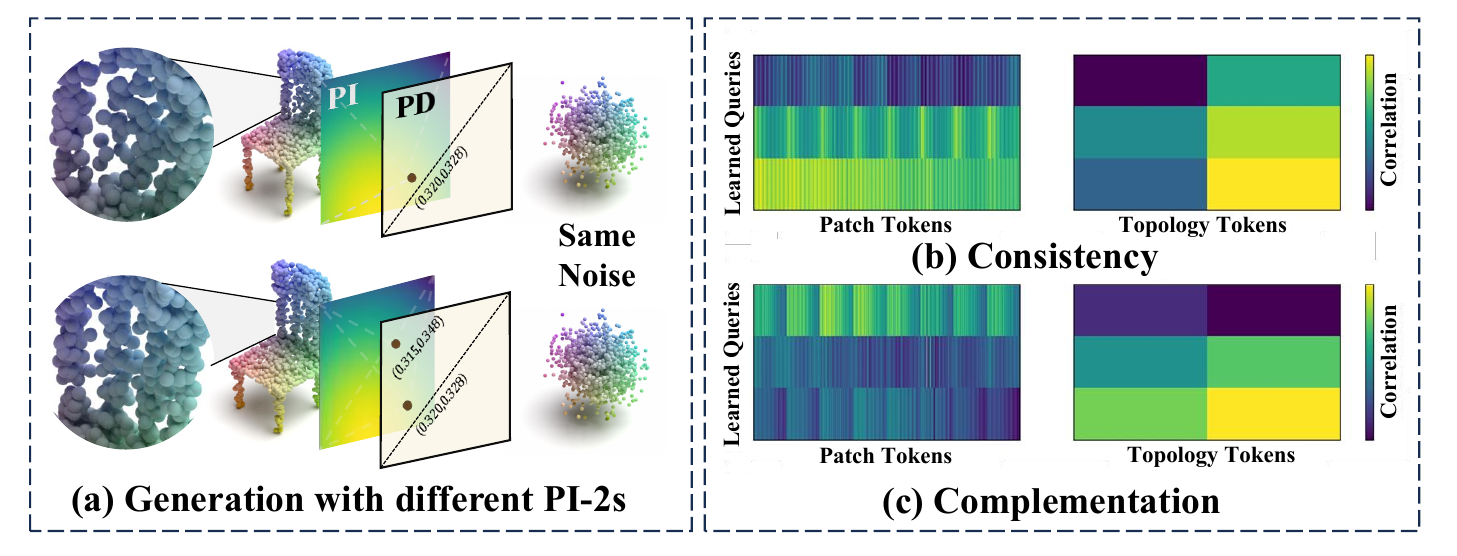} % Reduce the figure size so that it is slightly narrower than the column.
\vspace{-0.2cm}
\caption{The impact of topological information.}
\label{generative_process}
\end{figure}

\paragraph{Ablation studies on the impact of topological information.}
To evaluate the impact of topological information on point cloud generation, we input the same raw noise with different persistence images into the model to generate point clouds. As shown in Figure~\ref{generative_process}(a), richer topological information enhances fine-grained details in the generated point clouds, such as the voids in a chair’s backrest.

To further explore the synergy between topological and local geometric features, we compute the cosine similarity between learned latent queries and both patch tokens and topology tokens. Our finding reveals two distinct integration modes: \textbf{consistency} (Figure~\ref{generative_process}(b)), where a higher correlation with topology tokens aligns with a higher correlation with patch tokens, indicating the model uses topology tokens to integrate local information for global understanding; and \textbf{complementarity} (Figure~\ref{generative_process}(c)), where a higher correlation with topology tokens corresponds to a lower correlation with patch tokens, suggesting that the model directly extracts global structural information from topology tokens. These modes enable the model to adaptively activate queries from different perspectives, enhancing its ability to capture and process multi-scale information.

\paragraph{Generation of shapeNet’s mug and bottle categories.}
To further highlight the generation ability of TopoDiT-3D, we tested it on mug and bottle categories which have only 149 and 340 training samples, respectively. Despite limited training data, as shown in Figure~\ref{mug_bottle}, TopoDiT-3D outperforms DiT-3D and LION on both 1-NNA and COV, which not only reflects its capability to adapt to smaller dataset and more complex categories but also underscores its ability to generate high-quality 3D shapes, especially for intricate structures like mug openings and handles.

\paragraph{Scaling studies on the model sizes.}
As shown in Table~\ref{tab_model_size_ablation}, TopoDiT-3D significantly improves on 1-NNA and COV as the model scale increases, confirming that the bottleneck structure preserves the scalability of DiT. By better capturing complex structures and global information, starting from TopoDiT-3D-B/4, our model outperforms DiT-3D-XL/4. TopoDiT-3D-XL/4 achieves the best 1-NNA, while TopoDiT-3D-L/4 excels in COV. Notably, the VAE enhances persistence image diversity, improving both generation quality (lower 1-NNA) and diversity (higher COV).

\begin{table*}[t]
\centering
\begin{minipage}{0.48\textwidth}
\centering
\caption{Ablation on TopoDiT-3D model sizes for car generation.}
\label{tab_model_size_ablation}
\begin{adjustbox}{width=\textwidth}
\begin{tabular}{lllll}
\hline
\multicolumn{1}{c}{\multirow{2}{*}{Model}} & \multicolumn{2}{c}{1-NNA$(\downarrow)$}       & \multicolumn{2}{c}{COV ($\uparrow$)}         \\
\multicolumn{1}{c}{}                       & CD             & EMD            & CD             & EMD            \\ \hline
DiT-3D-XL/4                                & 48.24          & 49.35          & 50.00          & 56.38          \\ \hline
TopoDiT-3D-S/4                             & 57.95          & 59.09          & 55.68          & 51.98          \\
TopoDiT-3D-B/4                             & 45.31          & 43.89          & 55.39          & 58.23          \\
TopoDiT-3D-L/4                             & 44.48          & 44.88          & \textbf{58.52} & \textbf{63.06} \\
TopoDiT-3D-XL/4                            & \textbf{44.46} & \textbf{43.46} & 55.68          & 62.50          \\ \hline
\end{tabular}
\end{adjustbox}
\end{minipage}
\hfill
\begin{minipage}{0.44\textwidth}
\centering
\caption{Ablation on TopoDiT-3D resolution and downsampling.}
\label{tab_model_hyperparameters_ablation}
\begin{adjustbox}{width=\textwidth}
\begin{tabular}{lllllll}
\hline
\multicolumn{2}{c}{\multirow{2}{*}{\begin{tabular}[c]{@{}c@{}}Our Model\\ \color[HTML]{3166FF}{(Defalut:V=32,P=4,Q=96)}\end{tabular}}} & \multicolumn{2}{c}{1-NNA ($\downarrow$)} & \multicolumn{2}{c}{COV ($\uparrow$)} & \multirow{2}{*}{\begin{tabular}[c]{@{}l@{}}Train\\ Cost\end{tabular}} \\
\multicolumn{2}{c}{} & \multicolumn{1}{c}{CD} & \multicolumn{1}{c}{EMD} & \multicolumn{1}{c}{CD} & EMD &  \\ \hline
DiT-3D-S/4 & V=32,P=4 & 56.31 & 55.82 & 47.21 & 50.75 & 45 \\
Topo-3D-S/4 & Default & 49.84 & 46.46 & 58.28 & \textbf{60.69} & 30 \\ \hline
\multirow{2}{*}{Voxel Size} & V=16 & 49.02 & 47.36 & 55.87 & 57.68 & \textbf{26} \\
 & V=64 & \textbf{43.44} & \textbf{45.03} & 60.84 & 59.93 & 150 \\
\multirow{2}{*}{Patch Size} & P=2 & 49.32 & 50.97 & \textbf{61.59} & 60.09 & 150 \\
 & P=8 & 62.19 & 59.71 & 53.76 & 55.72 & \textbf{26} \\
\multirow{2}{*}{Learned Query} & Q=16 & 51.98 & 50.47 & 56.32 & 55.62 & 28 \\
 & Q=128 & 50.37 & 46.09 & 58.13 & 59.78 & 43 \\ \hline
\end{tabular}
\end{adjustbox}
\end{minipage}
\end{table*}

\paragraph{Ablation studies on the effects of voxel and patch sizes.}

We ablate the patch size from \{2, 4, 8\} and the voxel size from \{16, 32, 64\} to further explore the performance of the TopoDiT-3D model. As shown in table~\ref{tab_model_hyperparameters_ablation}, when the voxel size is set to 64, the model is able to capture more detailed information due to the higher resolution and achieve the best results. Similarly, a smaller patch size can reduce information loss. When the patch size is set to 2, the model shows the best performance in terms of CD. This trend is also noticeable in the DiT \cite{peebles2023DiT-2D} and DiT-3D \cite{mo2024dit} models. 

By using the Perceiver Resampler, the model efficiently represents voxel features with just 16 learned queries, even though 97\% of the tokens in the DiT block are reduced. Compared to the baseline model DiT-3D-S/4, TopoDiT-3D reduces the CD and EMD metrics by 4.33 and 5.35 on the 1-NNA and COV metrics, respectively, while improving CD and EMD by 9.11 and 9.03. This demonstrates that, despite the voxelization process generating numerous tokens, most of them do not carry significant information.

\begin{figure}[t]
\centering
\includegraphics[width=1\textwidth]{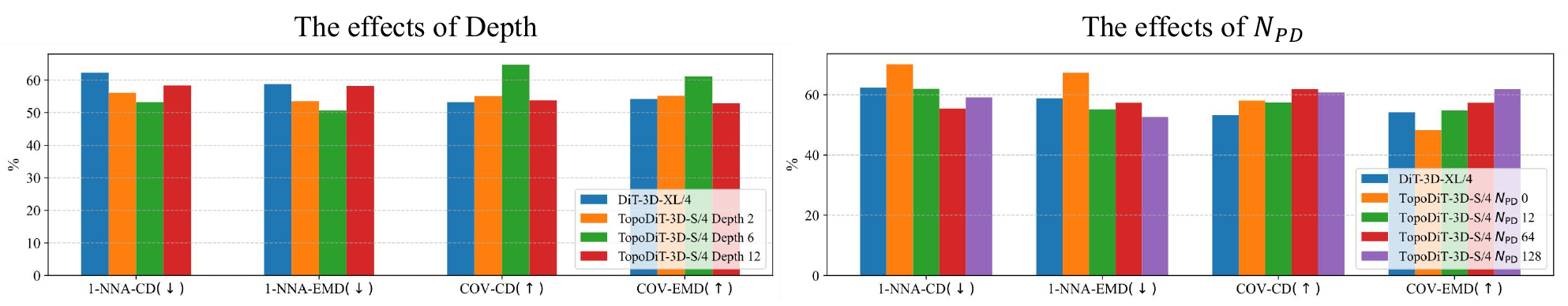} % Reduce the figure size so that it is slightly narrower than the column.
\caption{Ablation studies on the effects of depth of Perceiver Resampler for the airplane generation task.}
\label{ablation_depth}
\end{figure}

\paragraph{Ablation studies on the design of Perceiver Resampler and the number of key points $N_{\text{PD}}$.}
Figure~\ref{ablation_depth} presents experiments varying the Perceiver Resampler depth and the number of key points $N_{\text{PD}}$ between \{2, 6, 12\} and \{0, 12, 64, 128\}. Regardless of the Perceiver Resampler depth, the bottleneck structure not only effectively represents point-voxel features, but also integrates topological information to enhance the TopoDiT-3D-S/4 model performance, even exceeding the DiT-3D-XL/4 baseline. Without topological information ($N_{\text{PD}}$=0), TopoDiT-3D-S/4 marginally underperforms compared to DiT-3D-XL/4. However, with only $N_{\text{PD}}$=12, TopoDiT-3D-S/4 outperforms DiT-3D-XL/4. This demonstrates the crucial role of global information in 3D point cloud generation tasks.

\section{Conclusion}
In this work, we propose TopoDiT-3D, a novel topology-aware diffusion transformer with a bottleneck structure for 3D point cloud generation. TopoDiT-3D can not only adaptively filters out redundant tokens but also incorporates multi-scale topological information extracted through persistent homology into feature learning. Extensive experiments demonstrate that TopoDiT-3D outperforms state-of-the-art models in visual quality, diversity, and training efficiency. Our work validates not only the viability and effectiveness of the bottleneck structure in 3D diffusion transformers but also the significance of global topological information and its synergy with local geometric features, an aspect that has been overlooked in previous models.

{
\small
\bibliographystyle{unsrtnat}
\bibliography{main}
}

%%%%%%%%%%%%%%%%%%%%%%%%%%%%%%%%%%%%%%%%%%%%%%%%%%%%%%%%%%%%
\clearpage
\appendix
\section*{Appendix} 
In this appendix, we provide the following detailed material:
\begin{itemize}
   \item Section \ref{definitions} offers precise definitions of concepts related to persistent homology.

   \item Section \ref{analyses} presents comprehensive experimental analyses, including evaluations of the pretrained VAE for persistence images, assessments of TopoDiT-3D’s generation ability, computational cost analysis, and the effect of sampling steps.

   \item Section \ref{visualization} contains additional qualitative visualizations of 3D point cloud generation on the ShapeNet dataset.

   \item Section \ref{future} discusses the limitations of this work and outlines potential directions for future research.
\end{itemize}

% In this appendix, we provide the following detailed material: First, in Section \ref{definitions}, we offer the fundamentals of the DDPM generative model and precise definitions for the concepts related to persistent homology. Then, in Section \ref{analyses}, we present additional experimental analyses on the pretrained VAE for persistence images, TopoDiT-3D’s generation ability, computational cost analysis and the effect of sampling steps on performance. In Section \ref{visualization}, we include more qualitative visualizations of 3D point cloud generation for the ShapeNet dataset. Finally, in Section \ref{future}, we thoroughly discuss the limitations of this work and suggest directions for future research.

\section{Precise Definitions of Persistent Homology}
\label{definitions}

In this section, we introduce the precise definition of Persistent Homology, primarily referencing \cite{gabrielsson2020topologylayer,cohen2005stability,fugacci2016persistent,hensel2021survey}.

A cell complex $\mathcal{X}$ is a topological space that can be constructed by taking the union of a sequence of topological spaces such that each $k$-skeleton $\mathcal{X}_k$ is obtained by attaching some number of $k$-dimensional cells by their boundaries to the  $(k-1)$-skeleton $\mathcal{X}_{k-1}$ via continuous maps. Although cell complexes offer considerable flexibility in representing topological spaces, simplicial complexes are generally more favored in computational contexts due to their combinatorial characteristics.

\newtheorem{theorem}{Theorem}
\newtheorem{definition}[theorem]{Definition}

\begin{definition}
A k-dimensional simplex $(v_0,\dots,v_k)$ is the convex combination of $k + 1$ vertices, ${(v_0),\dots,(v_k)}$.
\end{definition}

Note that deleting one of the vertices $v_i$ from a $k$-dimensional simplex $(v_0,\dots,v_k)$ yields a $(k-1)$-dimensional simplex $(v_0,\dots,\hat{v}_i,\dots,v_k)$ which is determined by the remaining vertices and called the $i$-th face of $(v_0,\dots,v_k)$. Simplices are employed to represent a space, but must satisfy some additional constraints so that they form a simplicial complex.

\begin{definition}
A simplicial complex $\mathcal{X}$ is a set of simplices such that
\begin{enumerate}
    \item Every face of a simplex from $\mathcal{X}$ is also in $\mathcal{X}$.
    \item The non-empty intersection of any two simplices $\sigma_1,\sigma_2 \in \mathcal{X}$ is a face of both $\sigma_1$ and $\sigma_2$.
\end{enumerate}
\end{definition}

Homology is an algebraic invariant of a topological space, encoding a great deal of information while still being efficiently computable in many cases. To define homology, we first construct a chain group of $\mathcal{X}$ and the $k$-th chain group $C_k (\mathcal{X})$ is the freely generated group generated by $k$-dimensional simplices. In particular, there exists a boundary homomorphism $\partial_k : C_k (\mathcal{X}) \to C_{k-1} (\mathcal{X})$ which satisfy $\partial_{k-1} \circ \partial_k = 0$. Explicitly,
\begin{equation}
    \partial_k (v_0,\dots,v_k) = \sum_{i=0}^k (-1)^i (v_0,\dots,\hat{v}_i,\dots,v_k)
\end{equation}
where $\hat{v}_i$ indicates that the $i$-th vertex has been removed. As a notational convenience, we will use the same symbol for a $k$-cell $\sigma \in \mathcal{X}$ and the associated basis vector $\sigma \in C_k (\mathcal{X})$. The vector spaces $C_k (\mathcal{X})$ can be defined over any field, but for the purposes of determining kernels and images of maps exactly, we will restrict ourselves to working over the two element field $\mathbb{Z}/2\mathbb{Z}$. Homology in dimension $k$ is defined as the quotient vector space
\begin{equation}
    H_k(\mathcal{X})=\mathrm{ker} \partial_k / \mathrm{im} \partial_{k+1}
\end{equation}

\begin{figure*}[t]
\centering
\includegraphics[width=1\linewidth]{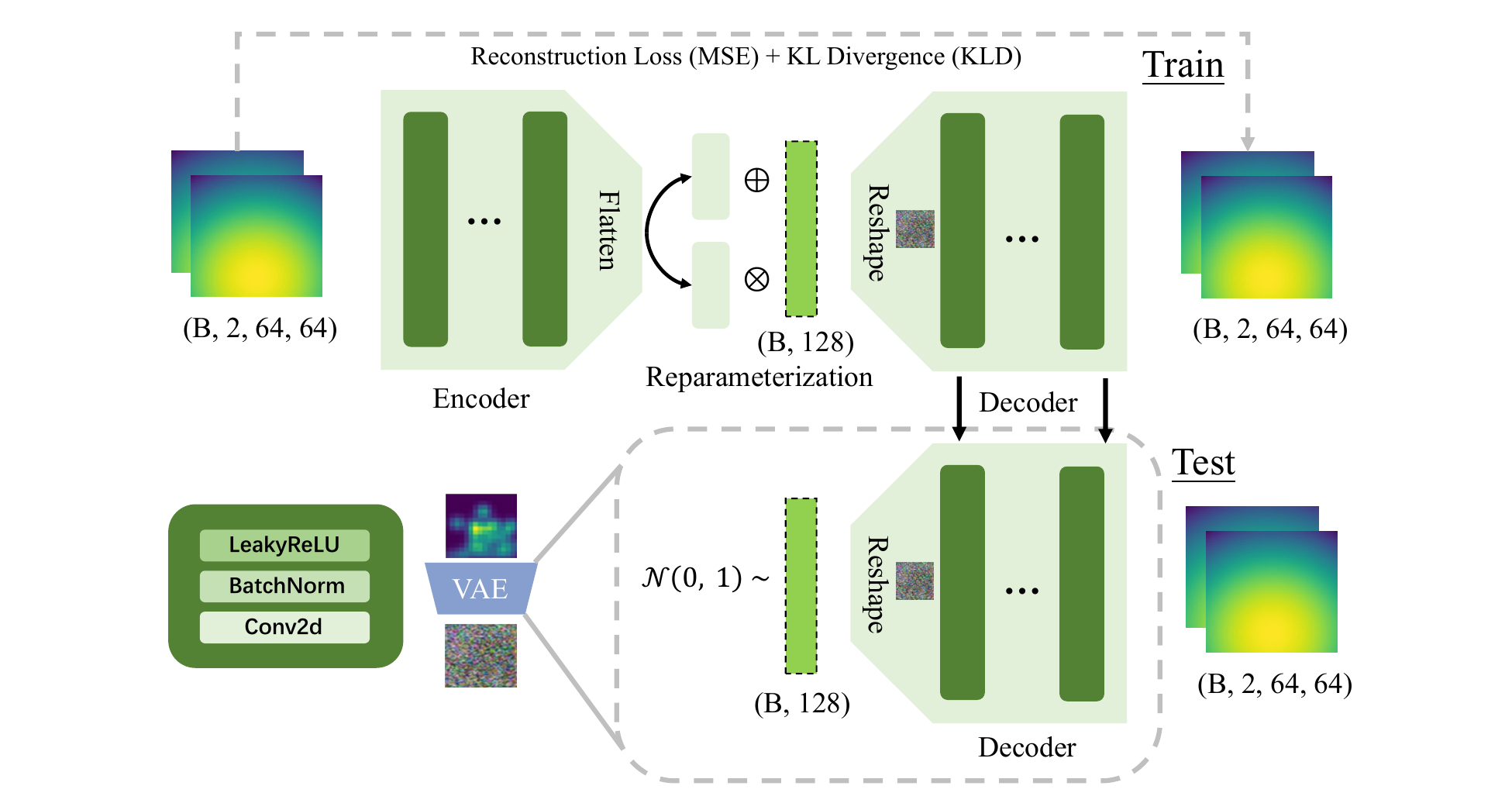} % Reduce the figure size so that it is slightly narrower than the column.
\caption{The architecture of VAE.}
\label{vae_arch}
\end{figure*}

\begin{figure*}[t]
\centering
\includegraphics[width=0.8\linewidth]{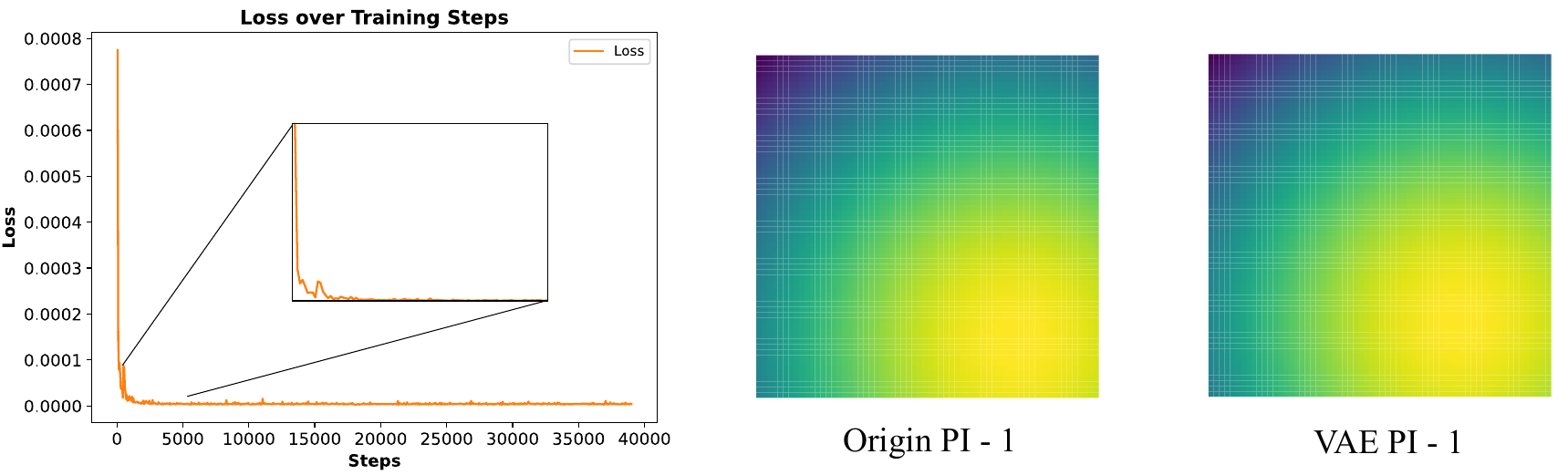} % Reduce the figure size so that it is slightly narrower than the column.
\caption{Loss over training steps of VAE, original PI and generated PI.}
\label{VAE}
\end{figure*}

An element of $H_k(\mathcal{X})$ is called a homology class, and a choice of representative for a class is called a generator. Homology is homotopy invariant, meaning that they remain invariant under homeomorphisms and therefore encode intrinsic information on the topology of $\mathcal{X}$. Consequently, homology and simpler invariants derived from them, such as the Betti-numbers $\beta_k :=\mathrm{dim}H_k(\mathcal{X})$, are valuable for counting the number of $k$-dimensional features of $\mathcal{X}$. For example, $\beta_0$ counts the number of connected components,  $\beta_1$ counts the number of holes, and so on. 

Persistent homology is an extension of homology to the setting of filtered chain complexes and serves as a key technique in topological data analysis (TDA). A filtered chain complex is a ascending sequence of chain complexes with inclusion maps. Filtered chain complexes naturally arise in situations where we have a sequence of inclusions of spaces $\mathcal{X}_0 \subseteq \mathcal{X}_1 \subseteq \dots \subseteq \mathcal{X}_n$. For example, we consider the sub-level sets $\mathcal{X}_\alpha = f^{-1}(-\infty,\alpha]$ of a so-called filtration function $f:\mathcal{X} \to \mathbb{R}$.

By employing a filtration, persistent homology can capture topological changes across different scales and encode this information in persistence diagrams. We consider a filtration where cells have a strict ordering, meaning they are added one at a time. Adding a $k$-dimensional cell $\sigma$ at parameter $i$ can lead to two outcomes. If $\partial_k \sigma$ is already in $\mathrm{im}\partial_k$ (meaning $\partial_k \sigma = \partial_k \omega$ for some $\omega \in C_k (\mathcal{X}_i \backslash \sigma)$), then $\omega - \sigma \in \mathrm{ker}\partial_k$. Since the kernel expands by one dimension, the quotient $H_k(\mathcal{X})=\mathrm{ker} \partial_k / \mathrm{im}\partial_{k+1}$ also expands by one dimension, with $\omega - \sigma$ generates the new homology class. Alternatively, if $\partial_k \sigma$ is not already in $\mathrm{im}\partial_k$, then $\mathrm{im}\partial_k$ expands by one dimension. Because $\partial_{k-1} \circ \partial_k=0$, $\partial_k \sigma \in \mathrm{ker}\partial_{k-1}$, and previously generated a homology class. Consequently, the quotient $H_{k-1}=\mathrm{ker} \partial_{k-1} / \mathrm{im}\partial_k$ will have one fewer dimension, and $\partial_k \sigma$ is a generator for the removed class. In summary, every cell in the filtration either creates or destroys homology when it appears. The complete details about how homology is born and dies throughout the filtration can be depicted as a multi-set of pairs $(b,d)$ where $b$ denotes the birth parameter of a homology class, while $d$ represents its death parameter. The persistence is given by $d-b$ and set to infinity if it never dies. This multi-set of pairs for homology in dimension $k$ is referred to the $k$-dimensional persistence diagram of the filtration, $\mathrm{PD}_k (\mathcal{X}_\alpha) = \{(b_i, d_i)\}_{i \in \mathcal{I}_k}$. 

The axes of the persistence diagram values at which topological features are created and destroyed, respectively. The single point with high persistence reflects the main topological characteristic of the point cloud, specifically its circular shape. Other topological features appear at smaller scales, resulting in a compact cluster in the lower-left area of the persistence diagram.

\section{Additional experimental analyses}
\label{analyses}
In this section, we conduct a series of additional experiments. These include evaluating the performance of the pretrained VAE for generating persistence images, performing an ablation study on the topological module, and assessing the generation ability of TopoDiT-3D. We also provide a detailed computational cost analysis of TopoDiT-3D. Additionally, we analyze how the number of sampling steps affects the model's performance.

\subsection{The pretrained VAE for persistence images}
To leverage these global features during the denoising process in inference, as illustrated in Figure~\ref{vae_arch}, we pretrain a VAE model to generate persistence images, guiding the model to generate point clouds. We visualize the loss variations of the VAE model during the training process for the chair category, as well as the comparison between the original PI and the generated PI by VAE, as shown in Figure~\ref{VAE}. During the training steps, the loss of the model decreases and approaches to 0, while the generated PI by VAE effectively reflects the topological structure as it is similar to the original PI, indicating that the model is capable of accurately reconstructing the input data and producing high-quality persistence images.

\subsection{Ablation study on the topological module}
\begin{figure}[t]
\centering
\includegraphics[width=0.6\linewidth]{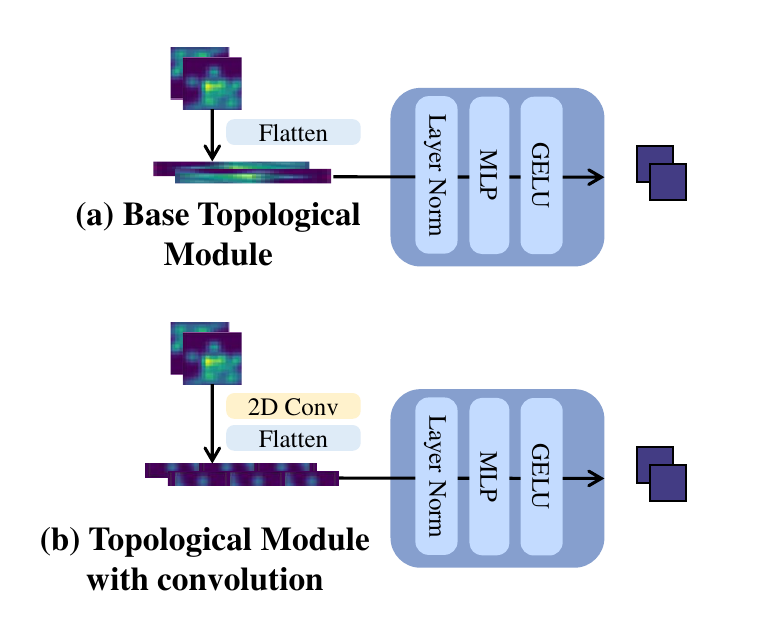} % Reduce the figure size so that it is slightly narrower than the column.
\caption{Different designs of the topological module for extracting Topology tokens.}
\label{Topological_Module_design}
\end{figure}

\begin{table}[t]
\caption{Exploration studies on different designs of the topological module for the chair generation task.}
\begin{adjustbox}{width=0.5\textwidth,center}
\begin{tabular}{lllll}
\hline
\multicolumn{1}{c}{\multirow{2}{*}{\begin{tabular}[c]{@{}c@{}}Method\\ (TopoDiT-3D-S/4)\end{tabular}}} & \multicolumn{2}{c}{1-NNA($\downarrow$)} & \multicolumn{2}{c}{COV($\uparrow$)} \\ 
\multicolumn{1}{c}{}                                                                                   & CD          & EMD         & CD         & EMD        \\ \hline
Base                                                                                                   & \textbf{52.86}       & 54.84       & \textbf{52.56}      & 51.05      \\
Base + Conv                                                                                            & 52.98       & \textbf{53.60}       & 51.71      & \textbf{52.56}      \\ \hline
\end{tabular}
\end{adjustbox}
\label{Topological_Module_result}
\end{table}

In Table~\ref{Topological_Module_result}, we employ a topological module with convolution, as depicted in Figure~\ref{Topological_Module_design}(b), to extract topology tokens from persistence images. The results demonstrate that both designs of the topological module achieve comparable performance. This can be attributed to the fact that persistence images derived from point clouds within the same category exhibit similar patterns due to their analogous topological structures. Therefore, our focus lies on the numerical variations between pixels in persistence images rather than low-level features such as edges or curves. The flattening operation does not compromise the integrity of global information, and the topological module effectively extracts topological information from persistent image encoding.

\subsection{TopoDiT-3D’s generation ability}
To demonstrate the generation ability of TopoDiT-3D across a broader range of categories, we conduct extensive testing on multiple classes and training datasets of varying sizes. In the following subsections, we detail TopoDiT-3D's performance on tasks involving sparse training data and multiple categories, highlighting its advantages in handling scarce data and complex structures.

\paragraph{TopoDiT-3D for other categories generation.}
\begin{figure*}[t]
\centering
\includegraphics[width=1\linewidth]{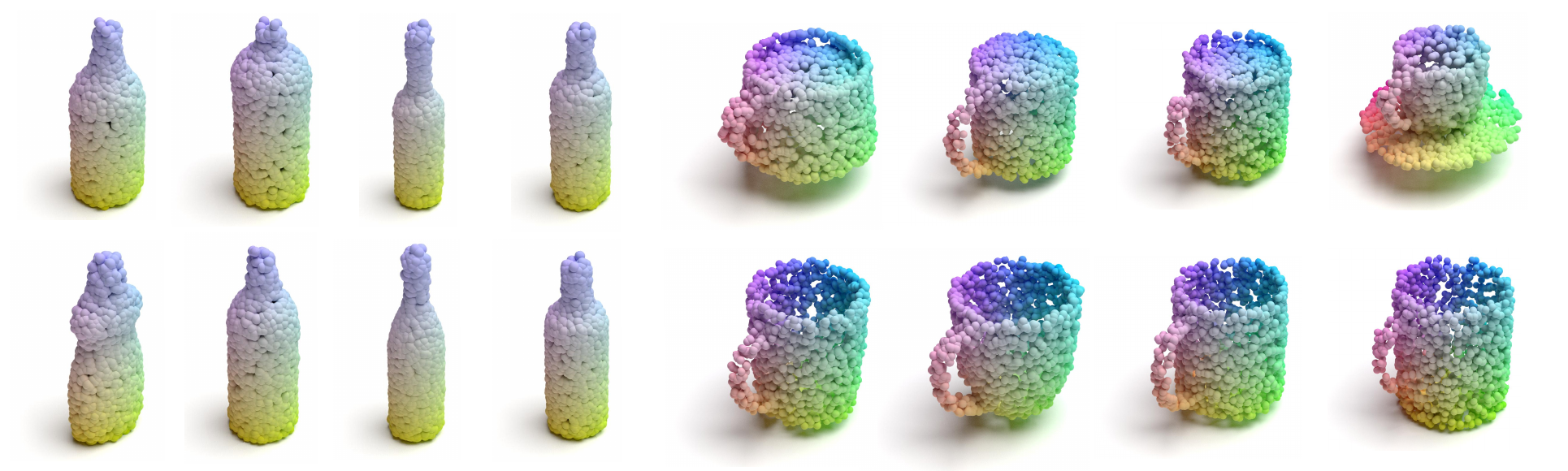} % Reduce the figure size so that it is slightly narrower than the column.
\caption{Qualitative visualizations on mug and bottle categories.}
\label{mug_bottle2}
\end{figure*}

To demonstrate the generation ability of TopoDiT-3D to a wider range of categories, we train it on the mug and bottle classes from ShapeNet, which contain only 149 and 340 training samples, respectively. These sample sizes are significantly smaller compared to common categories such as chair, airplane, and car, making it a challenging scenario for model generation. As shown in Table~\ref{mug_bottle_tab_appendix} and Figure~\ref{mug_bottle2}, TopoDiT-3D not only adapts successfully to these sparse training datasets but also outperforms DiT-3D in both the 1-NNA and COV evaluation metrics. This highlights the strong generation ability of TopoDiT-3D, as it effectively captures the topological features of mugs and bottles, thereby improving the quality of the generated 3D shapes. Moreover, TopoDiT-3D excels in generating complex structures, such as bottle openings and handles, further demonstrating its ability in handling diverse and intricate point cloud categories.

\paragraph{TopoDiT-3D for multi-categories generation and large-scale 55 categories generation.}

\begin{table*}[t]
\centering
\begin{minipage}{0.48\textwidth}
\caption{Exploration studies on mug and bottle categories generation.}
\label{mug_bottle_tab_appendix}
\centering
\begin{adjustbox}{width=\textwidth}
\begin{tabular}{llllll}
\hline
\multirow{2}{*}{Method} & \multirow{2}{*}{Category} & \multicolumn{2}{c}{1-NNA($\downarrow$)} & \multicolumn{2}{c}{COV($\uparrow$)} \\
 &  & CD & EMD & CD & EMD \\ \hline
DiT-3D-S/4 & Mug & 70.83 & 58.33 & 48.61 & 54.16 \\
TopoDiT-3D-S/4 & Mug & \textbf{59.03} & \textbf{53.47} & \textbf{50.00} & \textbf{59.72} \\ \hline
DiT-3D-S/4 & Bottle & 67.36 & 52.08 & 49.30 & \textbf{56.94} \\
TopoDiT-3D-S/4 & Bottle & \textbf{56.94} & \textbf{49.65} & \textbf{54.16} & 56.25 \\ \hline
\end{tabular}
\end{adjustbox}
\end{minipage}
\hfill
\begin{minipage}{0.48\textwidth}
\caption{Exploration studies on multi-class training.}
\label{table_multi_cate}
\centering
\begin{adjustbox}{width=\textwidth}
\begin{tabular}{lclllll}
\hline
\multirow{2}{*}{Model} & \multirow{2}{*}{\begin{tabular}[c]{@{}c@{}}Train\\ Class\end{tabular}} & \multirow{2}{*}{\begin{tabular}[c]{@{}l@{}}Test\\ Class\end{tabular}} & \multicolumn{2}{c}{1-NNA($\downarrow$)} & \multicolumn{2}{c}{COV($\uparrow$)} \\ \cline{4-7} 
& & & CD & EMD & CD & EMD \\ \hline
\multirow{3}{*}{DiT-3D-S/4} & Chair & Chair & 51.99 & 49.94 & 54.76 & 57.37 \\
& Chair,Car & Chair & 69.87 & 65.71 & 45.28 & 52.20 \\
& Chair,Car,Airplane & Chair & 75.80 & 74.72 & 41.46 & 47.89 \\ \hline
\multirow{3}{*}{TopoDiT-3D-S/4} & Chair & Chair & 49.84 & 46.46 & 58.28 & 60.69 \\
& Chair,Car & Chair & 60.46 & 60.31 & 56.17 & 56.92 \\
& Chair,Car,Airplane & Chair & 68.75 & 61.21 & 50.90 & 54.96 \\ \hline
\end{tabular}
\end{adjustbox}
\end{minipage}
\end{table*}

\begin{figure*}[t]
\centering
\includegraphics[width=0.85\linewidth]{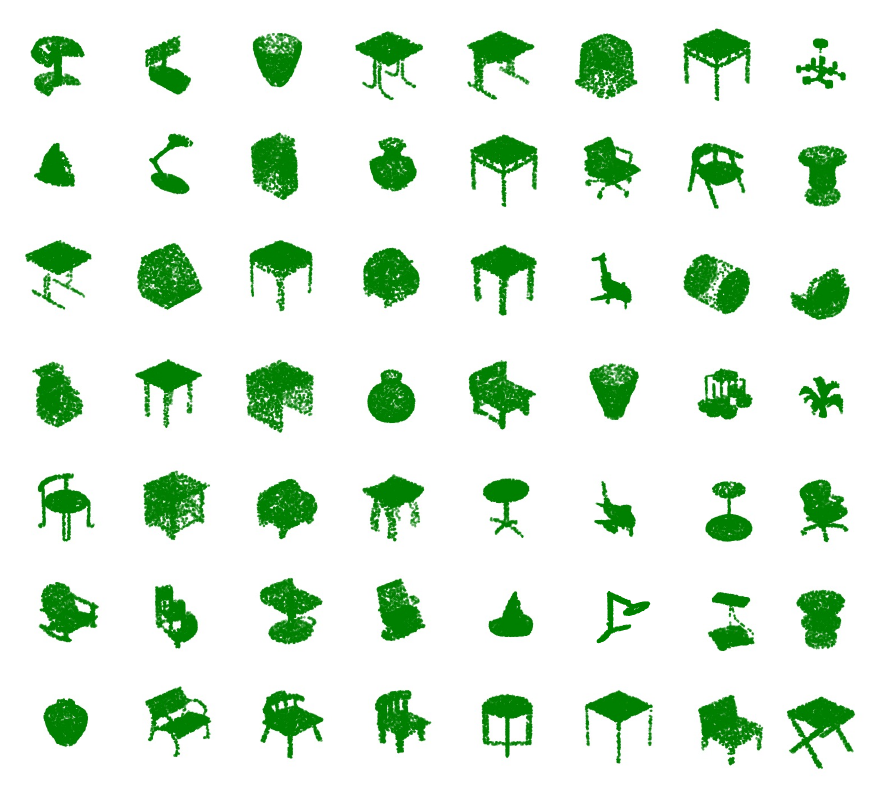} % Reduce the figure size so that it is slightly narrower than the column.
\caption{Qualitative visualizations on 55 different categories.}
\label{55categories}
\end{figure*}
To investigate the performance of our TopoDiT-3D on multi-class generation tasks, we chose \{Chair\}, \{Chair, Car\}, and \{Chair, Car, Airplane\} as training samples, respectively. We train TopoDiT-3D-S/4 and test its performance on the chair category, with the results shown in Table~\ref{table_multi_cate}. Compared to DiT-3D-S/4, our TopoDiT-3D-S/4 performs well across different category generation tasks due to the topological information of persistence images to help the model understand the underlying structure of random point clouds. We also train a TopoDiT-3D model on 55 different categories with 51k training samples \{airplane, bag, basket, bathtub, bed, bench, bottle, bowl, bus, cabinet, can, camera, cap, car, chair, clock, dishwasher, monitor, table and so on\}. The qualitative generation visualization is shown in Figure~\ref{55categories}.

\subsection{Computational cost analysis of TopoDiT-3D}
While integrating topological features enhances generation quality, it is essential to assess the computational cost. This section evaluates TopoDiT-3D's efficiency, comparing it to DiT-3D in terms of training time on different resolutions, and the cost of persistence diagram (PD) computation, persistence image (PI) extraction, and VAE training.

\begin{figure}[t]
\centering
\begin{minipage}[t]{0.38\linewidth}
    \centering
    \includegraphics[width=\linewidth]{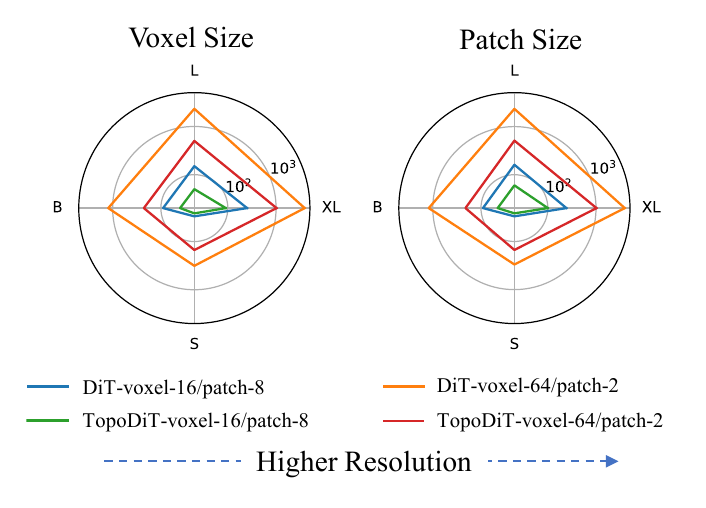}
    \vspace{-0.4cm}
    \caption{Computational cost across different resolutions.}
    \label{resolutions}
\end{minipage}
\hfill
\begin{minipage}[t]{0.58\linewidth}
    \centering
    \includegraphics[width=\linewidth]{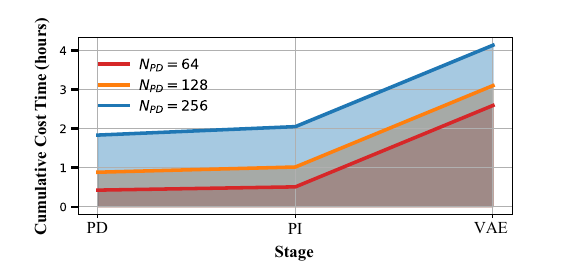}
    \vspace{-0.4cm}
    \caption{Analyzing computational cost across three stages: PD, PI data collection, and VAE training.}
    \label{PD/PI/VAE}
\end{minipage}
\end{figure}

\paragraph{Computational cost for large-scale deployment and higher voxel sizes.}
% \begin{figure}[t]
% \centering
% \includegraphics[width=1\linewidth]{Appendix/figures/computational_cost1.pdf} % Reduce the figure size so that it is slightly narrower than the column.
% \caption{Computational cost across different resolutions.}
% \label{resolutions}
% \end{figure}
To evaluate the computational efficiency of TopoDiT-3D under different voxel resolutions and large-scale training scenarios, we compare its performance with DiT-3D across various model scales. On the 55-category ShapeNet dataset with 51k training samples, TopoDiT-3D achieves a 3.1$\times$ speedup compared to DiT-3D. Additionally, as shown in Figure~\ref{resolutions}, TopoDiT-3D demonstrates improved efficiency across different model sizes and resolutions when tested on the chair category (5.2k training samples). This highlights its training efficiency and feasibility for large-scale 3D generation tasks.

\paragraph{Time for PD/PI collection and VAE training.}

% \begin{figure}[t]
% \centering
% \includegraphics[width=1\linewidth]{Appendix/figures/computational_cost2.pdf} % Reduce the figure size so that it is slightly narrower than the column.
% \caption{Analyzing computational cost cross three stages: PD, PI data collection, and VAE training.}
% \label{PD/PI/VAE}
% \end{figure}
One of the key components of TopoDiT-3D is the extraction of topological features through persistence diagrams (PD) and persistence images (PI). To optimize this step, we employ farthest point sampling (FPS) for point cloud $N_\textit{PD}$, which significantly reduces the computation cost of PD while preserving shape homology \cite{levrard2022distance}. As shown in Figure~\ref{PD/PI/VAE}, we analyze the computational cost at three stages: PD computation, PI data collection, and VAE training. The results indicate that under varying $N_\textit{PD}$ settings (tested on a single NVIDIA GeForce RTX 3090 GPU, batch size = 64, chair category with 5.2k samples), the overall time cost remains manageable. Furthermore, since VAE training and TopoDiT-3D training can be decoupled and executed in parallel, this provides an additional advantage in reducing training overhead. This parallelization capability allows for efficient large-scale training while maintaining high-quality topological feature learning.

\subsection{Effect of Sampling Steps}
\begin{figure*}[t]
\centering
\includegraphics[width=1\linewidth]{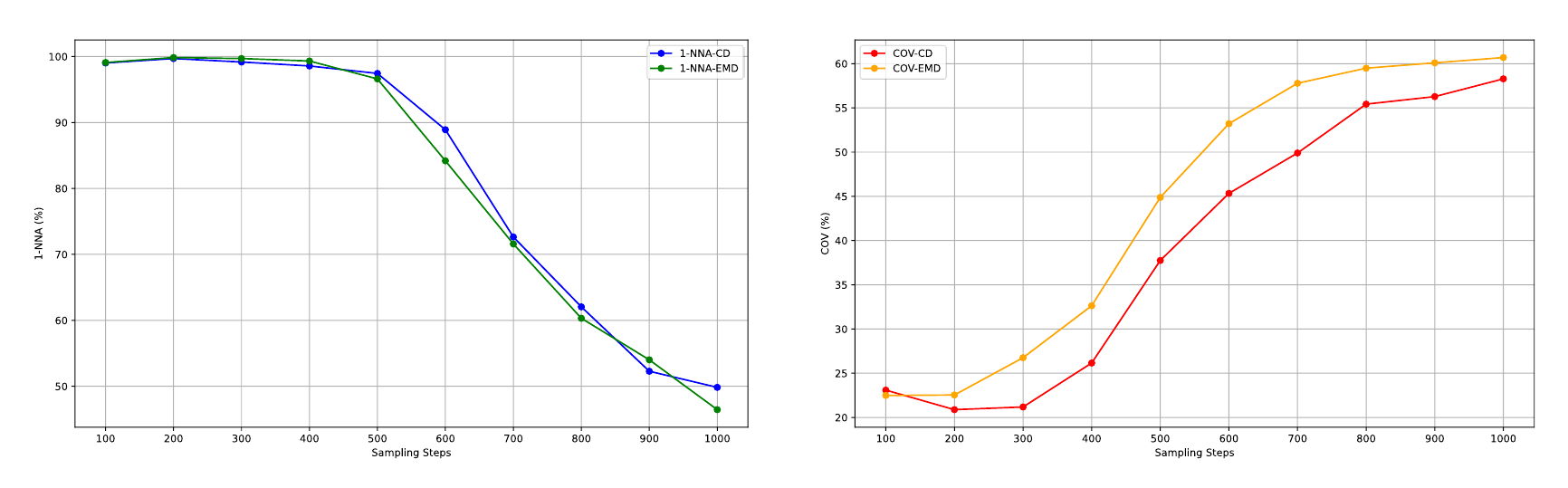} % Reduce the figure size so that it is slightly narrower than the column.
\caption{Effect of sampling steps on 3D shape generation (TopoDiT-3D-S/4, Chair) during the inference stage.}
\label{sampling_steps}
\end{figure*}

We explore the impact of the DDPM sampling step $T$ on the final performance during the inference stage in Figure~\ref{sampling_steps}. The proposed TopoDiT-3D obtains the best results (lowest 1-NNA and highest COV) for all metrics (CD and EMD) when the number of sampling steps is set to 1000. This trend is consistent with similar conclusions reached in previous DDPM work \cite{mo2024dit,vahdat2022lion}. Furthermore, when the batch size is set to 1 (i.e., generating a single point cloud sample) on an NVIDIA GeForce RTX 3090 with 1000 sampling steps, the generation speeds of TopoDiT-3D and DiT-3D are presented in Table~\ref{generation_speed}. It can be seen that TopoDiT-3D consistently achieves faster generation speeds compared to DiT-3D, regardless of model size is S/4 or XL/4, demonstrating its suitability for use on consumer-grade GPUs.

\begin{table}[t]
\centering
\caption{Generation speed of TopoDiT-3D and DiT-3D.}
\label{generation_speed}
\begin{adjustbox}{width=0.42\textwidth}
\begin{tabular}{lcc}
\hline
Model & Model Size & Speed (s/it) \\ \hline
\multirow{2}{*}{TopoDiT-3D} & S/4 & 8.20 \\
                            & XL/4 & 93.00 \\ \hline
\multirow{2}{*}{DiT-3D}     & S/4 & 8.50 \\
                            & XL/4 & 103.24 \\ \hline
\end{tabular}
\end{adjustbox}
\end{table}

\section{The qualitative generation visualization}
\label{visualization}

\begin{figure*}[!ht]
\centering
\includegraphics[width=0.78\linewidth]{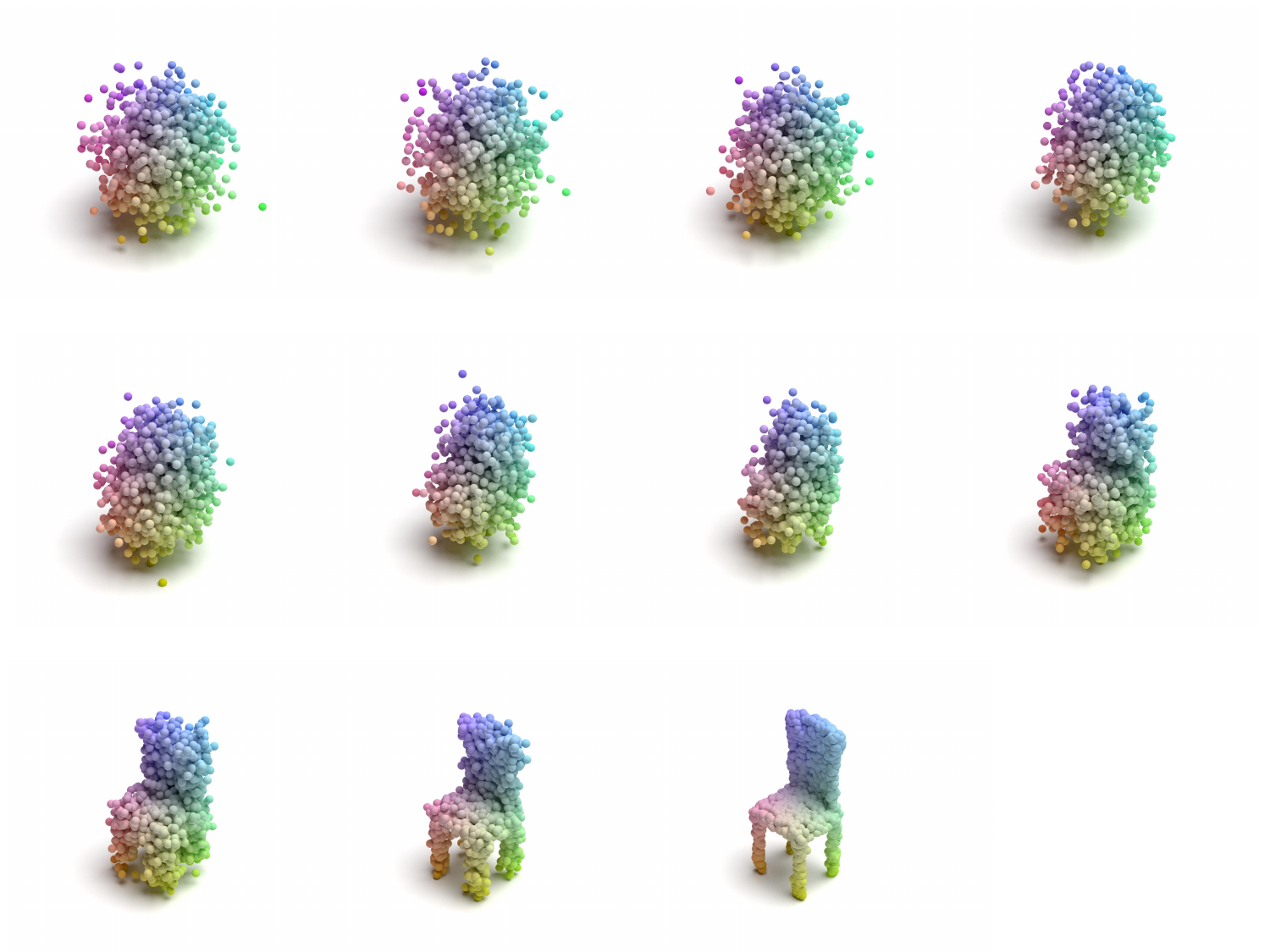} % Reduce the figure size so that it is slightly narrower than the column.
\caption{Qualitative visualizations of denoising process for chair generation.}
\label{chair_samples}
\end{figure*}

\begin{figure*}[!ht]
\centering
\includegraphics[width=0.78\linewidth]{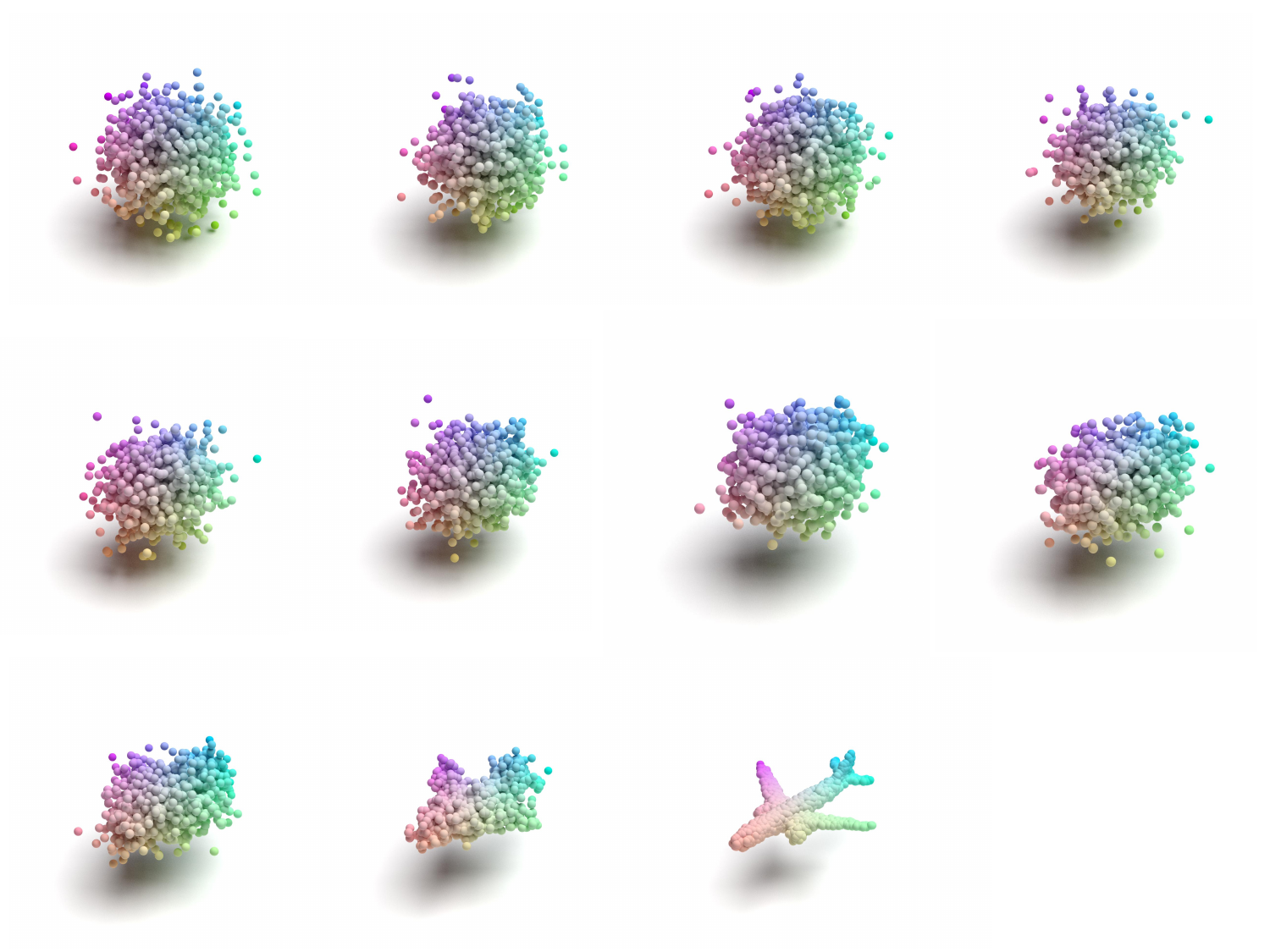} % Reduce the figure size so that it is slightly narrower than the column.
\caption{Qualitative visualizations of denoising process for airplane generation.}
\label{airplane_samples}
\end{figure*}

\begin{figure*}[!ht]
\centering
\includegraphics[width=0.78\linewidth]{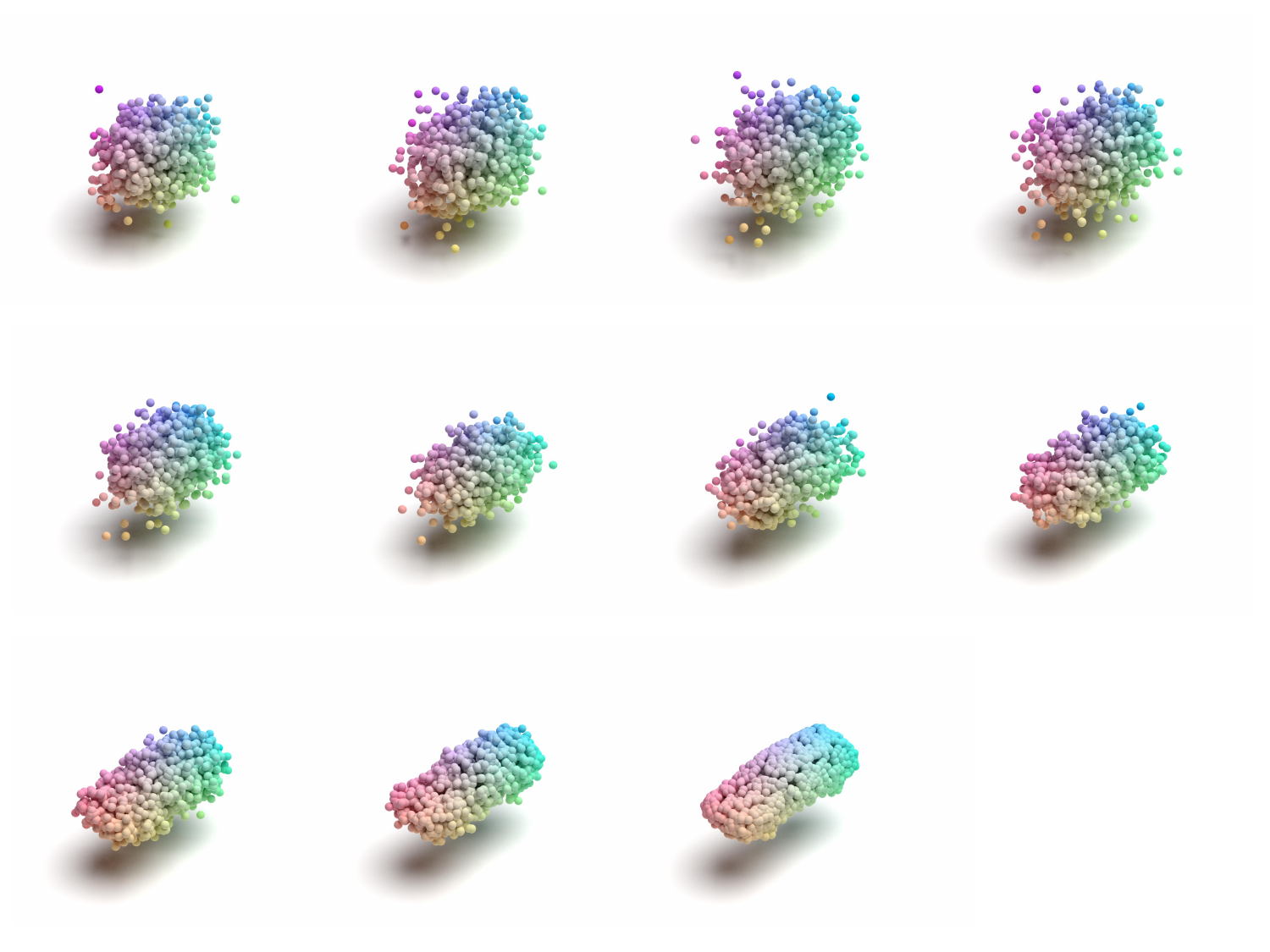} % Reduce the figure size so that it is slightly narrower than the column.
\caption{Qualitative visualizations of denoising process for car generation.}
\label{car_samples}
\end{figure*}

\begin{figure*}[!ht]
\centering
\includegraphics[width=1.0\linewidth]{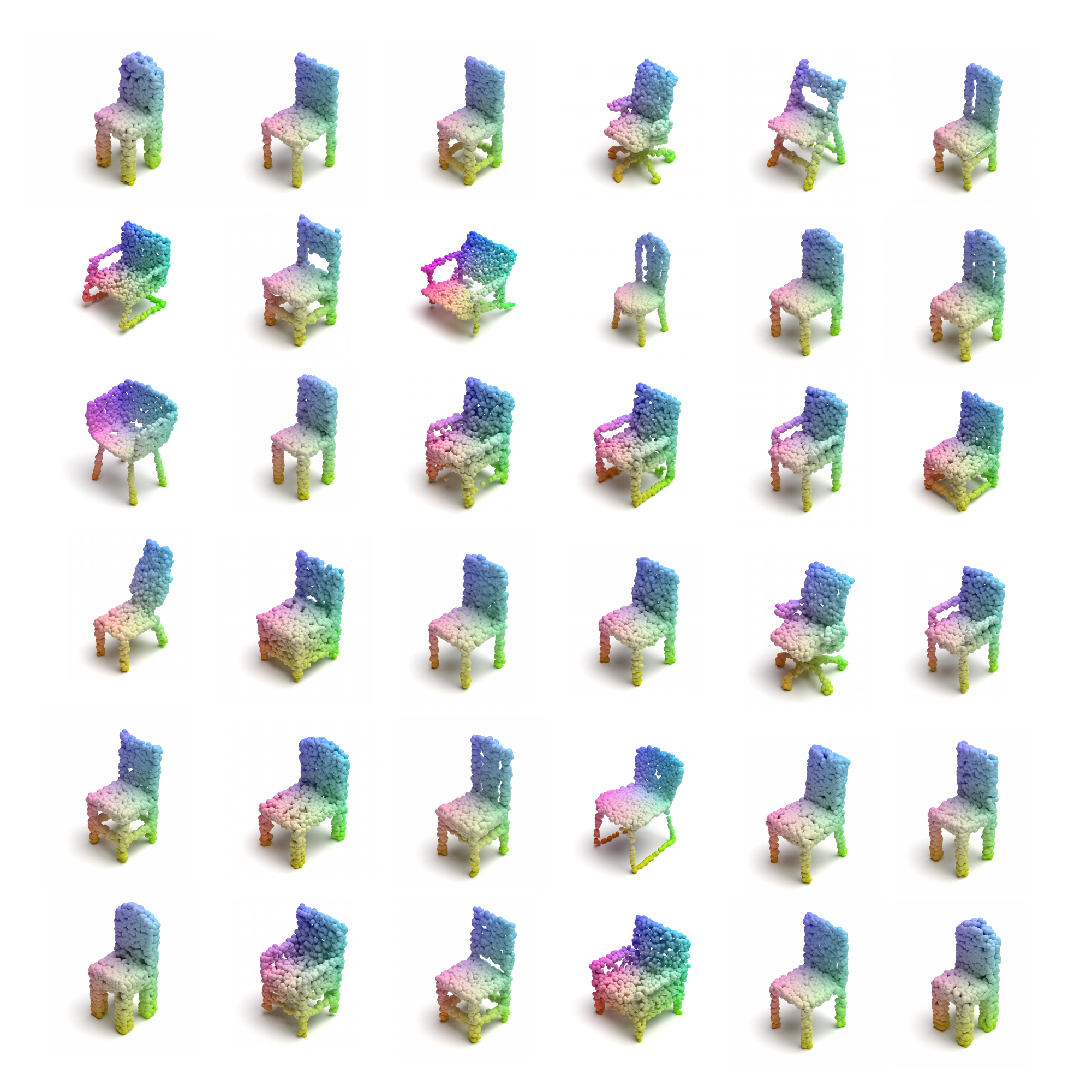} % Reduce the figure size so that it is slightly narrower than the column.
\caption{Qualitative visualizations of more generated shapes on chair category.}
\label{xr_chair}
\end{figure*}

\begin{figure*}[!ht]
\centering
\includegraphics[width=1.0\linewidth]{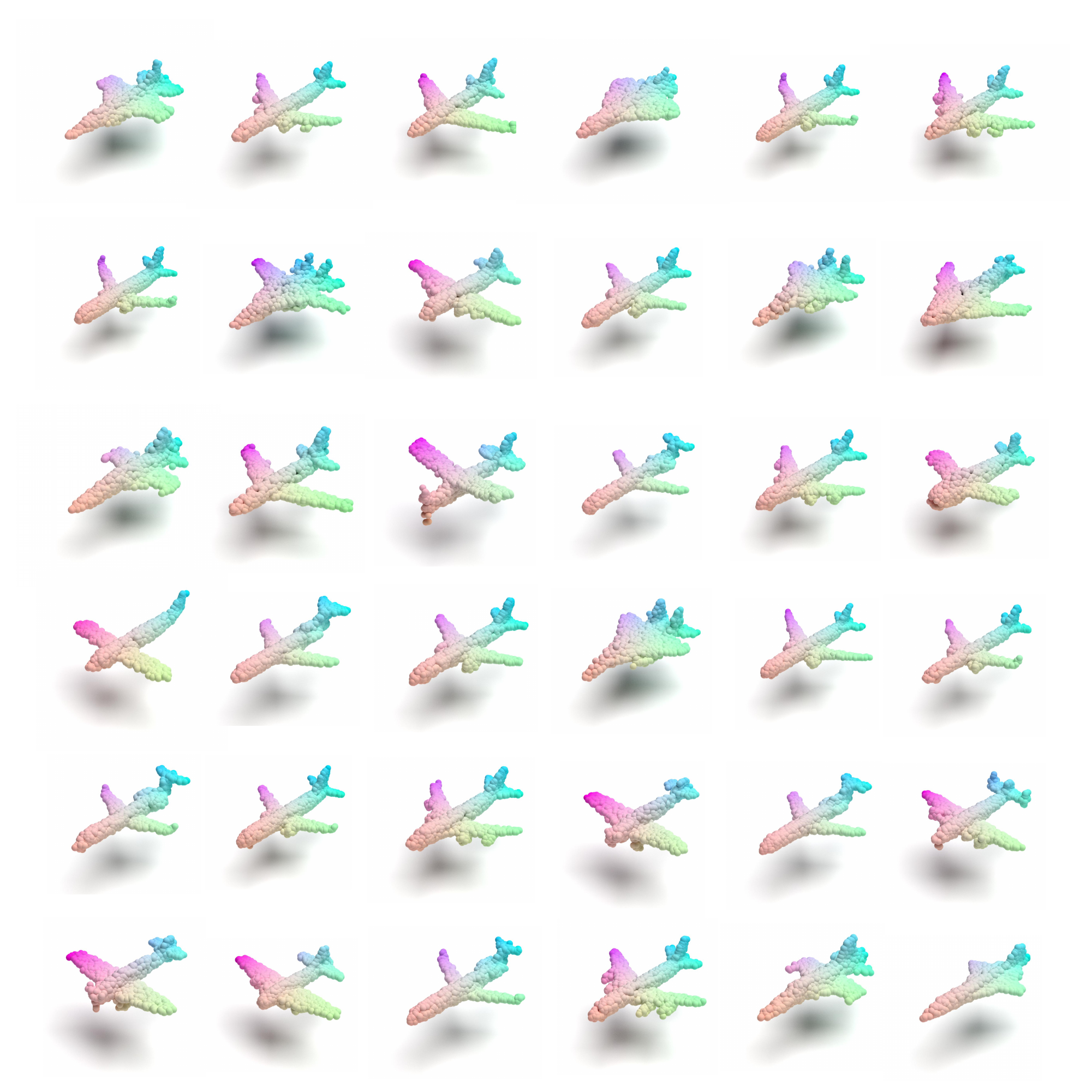} % Reduce the figure size so that it is slightly narrower than the column.
\caption{Qualitative visualizations of more generated shapes on airplane category.}
\label{xr_airplane}
\end{figure*}

\begin{figure*}[!ht]
\centering
\includegraphics[width=1.0\linewidth]{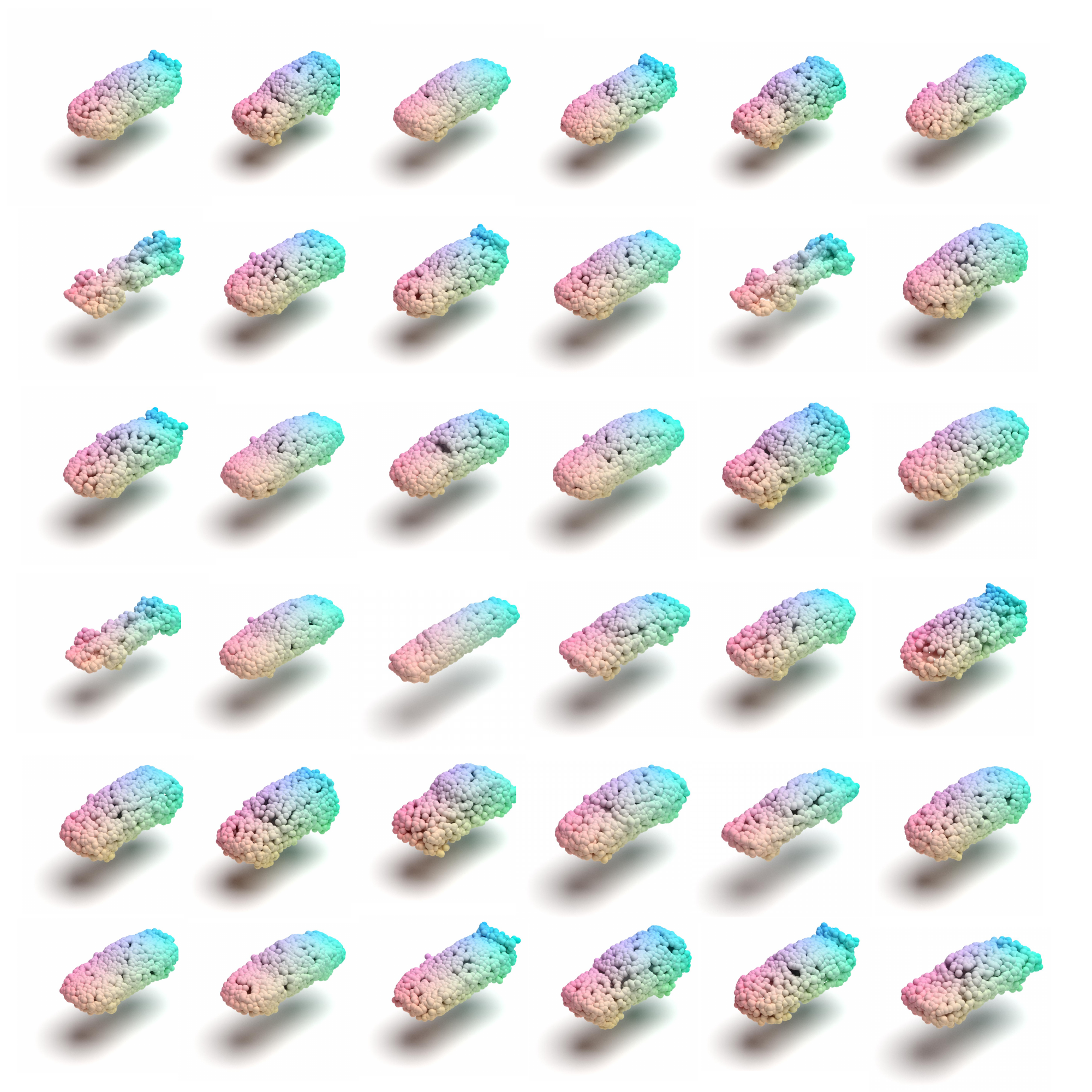} % Reduce the figure size so that it is slightly narrower than the column.
\caption{Qualitative visualizations of more generated shapes on car category.}
\label{xr_car}
\end{figure*}

In order to qualitatively demonstrate the effectiveness of the proposed TopoDiT-3D in 3D point clouds generation, we visualize the denoising process of chair, airplane, and car categories in Figures~\ref{chair_samples},~\ref{airplane_samples},~\ref{car_samples} and provide more visualization of 3D point clouds generated by our approach in Figures~\ref{xr_chair},~\ref{xr_airplane},~\ref{xr_car}.

\section{Discussion}
\label{future}

In this work, we unveil the crucial role of topological information in 3D point cloud generation and propose TopoDiT-3D, a novel topology-aware diffusion transformer bottleneck architecture. However, we have yet to investigate the impact of topological information on other 3D representations and dataset, such as meshes. Additionally, exploring the use of other complexes, such as the Alpha complex \cite{pun2022persistent}, for persistent homology could offer advantages in handling non-Euclidean data and potentially lead to more accurate topological descriptions. More interestingly, our work may offer a new thinking for controllable point cloud generation: by controllably generating persistence images, we can generate point clouds with specific topological structures. These directions hold great promise, and we will leave them as the future work.

% \begin{figure*}[!ht]
% \centering
% \includegraphics[width=0.78\linewidth]{nips_figures/car1.pdf} % Reduce the figure size so that it is slightly narrower than the column.
% \end{figure*}

%%%%%%%%%%%%%%%%%%%%%%%%%%%%%%%%%%%%%%%%%%%%%%%%%%%%%%%%%%%%

\end{document}